\documentclass{article} % For LaTeX2e
\usepackage{arxiv,times}
\usepackage{xcolor}
\definecolor{lightgray}{gray}{0.9}

\usepackage{amsmath,amsfonts,bm}

\def\eqref#1{equation~\ref{#1}}
\def\1{\bm{1}}

\newcommand{\brac}[1]{\left(#1\right)}
\newcommand{\sbrac}[1]{\left[#1\right]}

\DeclareMathAlphabet{\mathsfit}{\encodingdefault}{\sfdefault}{m}{sl}
\SetMathAlphabet{\mathsfit}{bold}{\encodingdefault}{\sfdefault}{bx}{n}

\def\tX{{\tilde{X}}}

\def\gX{{\mathcal{X}}}
\def\gY{{\mathcal{Y}}}
\def\gZ{{\mathcal{Z}}}

\def\sR{{\mathbb{R}}}

\newcommand{\E}{\mathbb{E}}
\newcommand{\Ls}{\mathcal{L}}

\newcommand{\KL}{D_{\mathrm{KL}}}

\DeclareMathOperator*{\argmax}{arg\,max}

\usepackage{algorithm}
\usepackage{algpseudocode}
\usepackage{hyperref}
\usepackage{booktabs}
\usepackage{graphicx}
\usepackage{subcaption}
\usepackage{url}
\usepackage{multirow}
\usepackage{bbm}
\usepackage{amsthm}
\usepackage{enumitem}
\usepackage{wrapfig}
\usepackage[subtle]{savetrees}
\newtheorem{theorem}{Theorem}
\newtheorem{lemma}{Lemma}

\title{
    Towards Unsupervised Speech Recognition\\
    at the Syllable-Level
}

\author{Liming Wang$^1$,
Junrui Ni$^2$\thanks{Work done at UIUC, now at Amazon}, Kai-Wei Chang$^1$, Saurabhchand Bhati$^1$, David Harwath$^3$,\\ 
\textbf{Mark Hasegawa-Johnson}$^2$, \textbf{James R. Glass}$^1$\\
$^1$Massachusetts Institute of Technology, $^2$University of Illinois Urbana-Champaign,\\
$^3$University of Texas at Austin\\
\texttt{limingw@csail.mit.edu}
}

\newcommand\liming[1]{}
\newcommand{\dd}{\mathrm{d}}

\def\aishell/{AISHELL-3}
\def\hubl/{HuBERT}
\def\mincut/{min-cut}
\def\km/{K-means}
\def\reborn/{REBORN}
\def\sylcipher/{SylCipher}
\def\pdf/{probability distribution}
\def\prenet/{pre-net}
\def\postnet/{post-net}
\def\pyphen/{Pyphen}
\def\wtov/{wav2vec 2.0}
\def\wtovu/{wav2vec-U}
\def\secvsp{\vspace{-0.25cm}}
\def\subsecvsp{\vspace{-0.2cm}}
\def\figvsp{\vspace{-0.25cm}}

\iclrfinalcopy % Uncomment for camera-ready version, but NOT for submission.
\begin{document}

\maketitle
%\vspace*{-1.6cm}

\begin{abstract}
\secvsp
% Training speech recognizers with unpaired speech and text—commonly referred to as unsupervised speech recognition (UASR)—is a crucial step toward extending ASR to low-resource languages in the long-tail distribution and enabling multimodal learning from non-parallel data. However, existing phone-based approaches often depend on costly resources such as grapheme-to-phoneme converters (G2Ps) and suffer from poor generalization in languages with ambiguous phoneme boundaries, partly due to training instability. In this paper, we address both challenges by proposing a syllable-level UASR framework based on masked language modeling, which eliminates the reliance on G2Ps and avoids the instability of GAN-based methods. Our approach achieves up to a 40% relative reduction in character error rate (CER) on LibriSpeech and generalizes effectively to Mandarin, a language that has remained particularly difficult for prior methods.
Training speech recognizers with unpaired speech and text -- known as unsupervised speech recognition (UASR) -- is a crucial step toward extending ASR to low-resource languages in the long-tail distribution and enabling multimodal learning from non-parallel data. However, existing approaches based on phones often rely on costly resources such as grapheme-to-phoneme converters (G2Ps) and struggle to generalize to languages with ambiguous phoneme boundaries due to training instability. In this paper, we address both challenges by introducing a syllable-level UASR framework based on masked language modeling, which avoids the need for G2P and the instability of GAN-based methods. Our approach achieves up to a 40\% relative reduction in character error rate (CER) on LibriSpeech and generalizes effectively to Mandarin, a language that has remained particularly difficult for prior methods. Code will be released upon acceptance.\liming{create a github}
\end{abstract}
\secvsp

\section{Introduction\liming{(READY)}}\label{sec:intro}
\secvsp

% \liming{Universal $->$ pushing the horizon/long tail distribution of languages}
% \liming{Toward unsupervised syllabic speech recognition? Footnote? A more general intro on unpaired multimodal learning?}
% Why unsupervised speech recognition
Recent advances in self-supervised learning~\citep{Baevski2020-wav2vec2,Hsu2022-hubert,chen2022-wavlm,chung2021w2vbert,chen2024-xeus,mohamed2022self} and spoken language modeling~\citep{arora2025-slm,chu2024qwen2,defossez2024moshi} have enabled voice assistants with increasingly human-like listening and speaking abilities. However, they are far from \emph{language-universal}: most systems support only a handful of languages in the world~\citep{chen2024-xeus,conneau21_interspeech}, due to a lack of large-scale paired speech and text training corpora. 

A promising step toward language-universal assistants is to build speech recognizers from \emph{unpaired} speech and text, or \emph{unsupervised speech recognition} (UASR)~\citep{Glass2012-unsup-speech,Baevski2021-wav2vec-u,Liu2023-wav2vecu2,Tseng2024-reborn}. 
% Fundamental significance
UASR is a fundamental challenge: success would enable downstream tasks such as speech synthesis~\citep{Ni-unsuptts-interspeech2022,Liu2022-utts}, translation~\citep{wang-etal-2023-simple}, and understanding~\citep{shi2023-unsupslu}, paving the way for general-purpose voice assistants. It is also a central case of \emph{unpaired multimodal learning}~\citep{Artetxe18-unsupnmt,artetxe2018unsupervised,artetxe-etal-2019-effective,Lample2018-unsupmtmono,lample2018phrase,ma2019-unpaired,hoshen2018non}, where modalities need to be aligned without parallel data. In the absence of sentence-level alignment, the model must infer higher-level linguistic units--phones, syllables, and words--from raw speech waveforms in conjunction with global text statistics. Thus, UASR provides broader insights into representation learning and multimodal alignment without supervision.

% Why syllable-level
The best existing UASR systems~\citep{Baevski2021-wav2vec-u,Liu2023-wav2vecu2,Tseng2024-reborn} operate at the \emph{phoneme-level}. To this end, they need to convert raw text units, or \emph{graphemes}, to phonemes, or minimal sound units that encode meaning, using a grapheme-to-phoneme convertor (G2P). However, training G2Ps requires resources such as pronunciation dictionaries, which can be time-consuming and labor-intensive to create. Without a G2P, such systems suffer from significant performance degradation due to misalignment between speech and raw text in many languages~\cite{Liu2023-wav2vecu2,Ni-unsuptts-interspeech2022}. Even when pronunciation dictionaries are available, the system may fail to detect clear phone-level boundaries due to strong co-articulation effects for languages such as Mandarin.

An alternative approach to phone-based models is to build a word-level UASR system, which can be achieved without a G2P. Yet a major concern is the coverage of the system on \emph{rare words}, which can effectively be infinite in vocabulary size, making it significantly harder to acquire than phones.  Furthermore, detecting word boundaries requires capturing long-range contextual dependencies in speech, which risk destabilizing segmentation mechanisms effective for phone-level UASR~\cite{wang-etal-2023-unsup-speech2sign}.

In this work, we propose a third alternative -- to build UASR at the \emph{syllable-level}, which can be justified on three grounds. First, unlike words, the number of distinct syllables for a language is finite, which reduces the long-tail token distribution issue and allows for better generalization to unseen words; second, many languages exhibit the best alignment between speech and text at the syllable level instead of the phone or word level. 
% For instance, \emph{tonal} languages like Mandarin use tones to distinguish meaning, which are suprasegmental features realized at the syllable level, making the syllable more integrated as a whole rather than separate phonemes\liming{cite?}. 
For instance, Mandarin uses characters, which have strong correspondences to spoken syllables; therefore, we expect that a syllable-level UASR system could be more appropriate for UASR than a phoneme-based system, which we also found to be the case empirically.

Last but not least, recent advancement in syllable boundary detection and unit discovery~\citep{Baade2025-syllablelm,Cho2025-sylber} has made it possible to segment raw speech into syllable-like units without any textual supervision, and is often more reliable than unsupervised segmentation methods at the word-level, while capturing most of the word-level semantics~\cite{peng2022-vghubert,fuchs2023-gradseg}.

In this paper, we make the following contributions.
\begin{enumerate}[leftmargin=2em, labelsep=0.5em, itemindent=0.0em]
    \item This paper introduces \textbf{\sylcipher/}, to our knowledge, the first syllable-based UASR system. \sylcipher/ jointly predicts syllable boundaries and embedding tokens from raw speech using a unified self-supervised objective.  The learning mechanism avoids adversarial training, making it more stable and less sensitive to hyperparameters. 
    \item We provide an information-theoretic analysis on the proposed UASR system proving that our training objective achieves perfect distribution matching and zero-error UASR under regularity conditions.
    \item We conduct extensive experiments across domains and languages. On LibriSpeech, \sylcipher/ achieves up to 40\% relative character error rate (CER) reduction over prior G2P-free UASR methods. On SpokenCOCO, improvements are even larger, demonstrating robustness across domains. On Mandarin, \sylcipher/ achieves 12.2\% phone error rate (PER), outperforming GAN-based UASR methods that fail to even converge.
    \item We perform careful ablation and error analysis, examining the effects of token vocabulary size, syllabifier choice, and segmentation mechanisms.
\end{enumerate}
\textbf{Paper organizations.} Section~\ref{sec:related_work} reviews related work. Section~\ref{sec:formulation} formalizes the syllable-level UASR problem. Section~\ref{sec:method} presents \sylcipher/ and its theoretical guarantees. Section~\ref{sec:exp} reports experiments and ablations. Section~\ref{sec:concl} concludes with limitations and future work. 

\secvsp
\section{Related work\liming{(READY)}}\label{sec:related_work}
\secvsp
% UASR
\paragraph{Unsupervised speech recognition} Early work on UASR assumed the existence of a reliable G2P and formulate the problem as an adversarial game, where a conditional generator predicts a phoneme sequence given a speech waveform, and a discriminator tries to tell real phonemized text apart from the generator's output~\citep{wang-etal-2023-unsupasr-theory}. Within this framework, several works have explored different segmentation mechanisms such as fixed unsupervised phoneme segmentation~\citep{Liu2018-asru}, iterative forced alignment~\citep{Chen2019-uasr}, de-duplication~\cite{Baevski2021-wav2vec-u} and reinforcement learning~\citep{Tseng2024-reborn}. \citep{wang-etal-2023-unsup-speech2sign} proposed an alternative formulation of UASR as an explicit distribution matching problem, by matching the lower-order $N$-gram distributions of the generated and real texts. Later work further extended this reformulation to use G2P-free tokens such as words~\citep{wang-etal-2023-unsup-speech2sign,Ni2025-jstti}, more expressive architecture such as transformers~\citep{Ni2025-jstti} and more general text distribution such as masked prediction probabilities~\citep{Ni2025-jstti}. We adapt this word-level UASR approach to the syllable-level, and supplement it with a simplified training flow, a more stable differentiable boundary detector and additional learning objectives.
% Syllable-level SSL
\paragraph{Syllable-level self-supervised learning} Early work on syllable-level modeling of speech used signal processing techniques~\citep{Zhang2009-syllable}. To discover higher-level structure and more efficient self-supervised learning (SSL) representations from raw speech, \citep{Peng2023-syllable,cho2023-sdhubert} proposed to induce syllabic structure from existing SSL models such as HuBERT~\citep{Hsu2022-hubert}. To this end,~\citep{Peng2023-syllable} probed the self-attention layers of VG-HuBERT~\citep{peng2022-vghubert}, a visually grounded SSL model to detect syllable-like feature clusters and further refine such clusters using a \mincut/ algorithm~\cite{shi1997normalized}. To sidestep the need for visual data, \citep{cho2023-sdhubert} employed a speech-only SSL model trained with utterance-level self-distillation from a HuBERT teacher. Due to the indirect manner of such approaches by which syllabic structures are derived, they are often noisy and unreliable. To cope with this issue, recent works~\citep{Baade2025-syllablelm,Cho2025-sylber} proposed a more direct and targeted approach by performing self-distillation at the syllable-level, which significantly improved syllable boundary detection and unit discovery performance by encouraging sharper contrast between within and between-syllable feature frames. 

\secvsp
\section{Syllable-level unsupervised speech recognition\liming{(READY)}}\label{sec:formulation}
\secvsp
In this section, we formulate syllable-level UASR as follows. 
Let $X=[X_1,\cdots,X_T]\in\gX^T$ be a padded sequence of speech feature vectors and let $Y=[Y_1,\cdots,Y_L]\in\gY^L$ a padded sequence of text tokens in the same language. Since a tokenized speech utterance typically uses more tokens than the text transcription of the same utterance, we assume the $T\geq L$. Assume $X$ and $Y$ come from two \emph{unpaired} datasets and are therefore \emph{statistically independent}. Further, suppose they are \emph{matched}, i.e., there exists an ASR function $y^*:\gX\mapsto\gY$ such that the distributions of $X$ and $Y$, $p_X$ and $p_Y$ satisfy
\begin{align}
    p_Y(y)=\int_{x\in\gX:y^*(x)=y} p_X(x)\dd x,\,\forall y\in\gY^L, \label{eq:matched}
\end{align}
The goal of UASR is to recover $y^*$ given only unpaired $X$ and $Y$. The syllable-level case is a special setting where the tokens of $Y$ are syllables. The learning problem resembles \emph{decipherment}, where one decodes a message in an unknown script without a lexicon or grammar. In practice, $X$ is formed from frame-level SSL features~\cite{Hsu2022-hubert}, and $p_X$ and $p_Y$ are only approximately matched due to finite-sample noise and domain mismatch. As shown in \citep{wang-etal-2023-unsupasr-theory}, UASR is ill-posed in general, but the mapping $y^*$ in \eqref{eq:matched} becomes identifiable if syllable boundaries are known and the language satisfies mild conditions.

\secvsp
\section{SylCipher: Syllable-level UASR via information-constrained masked language modeling\liming{(READY)}}\label{sec:method}
\secvsp
\begin{figure}
    \centering
    \begin{subfigure}{0.63\textwidth}
        \centering
        \includegraphics[width=0.9\textwidth]{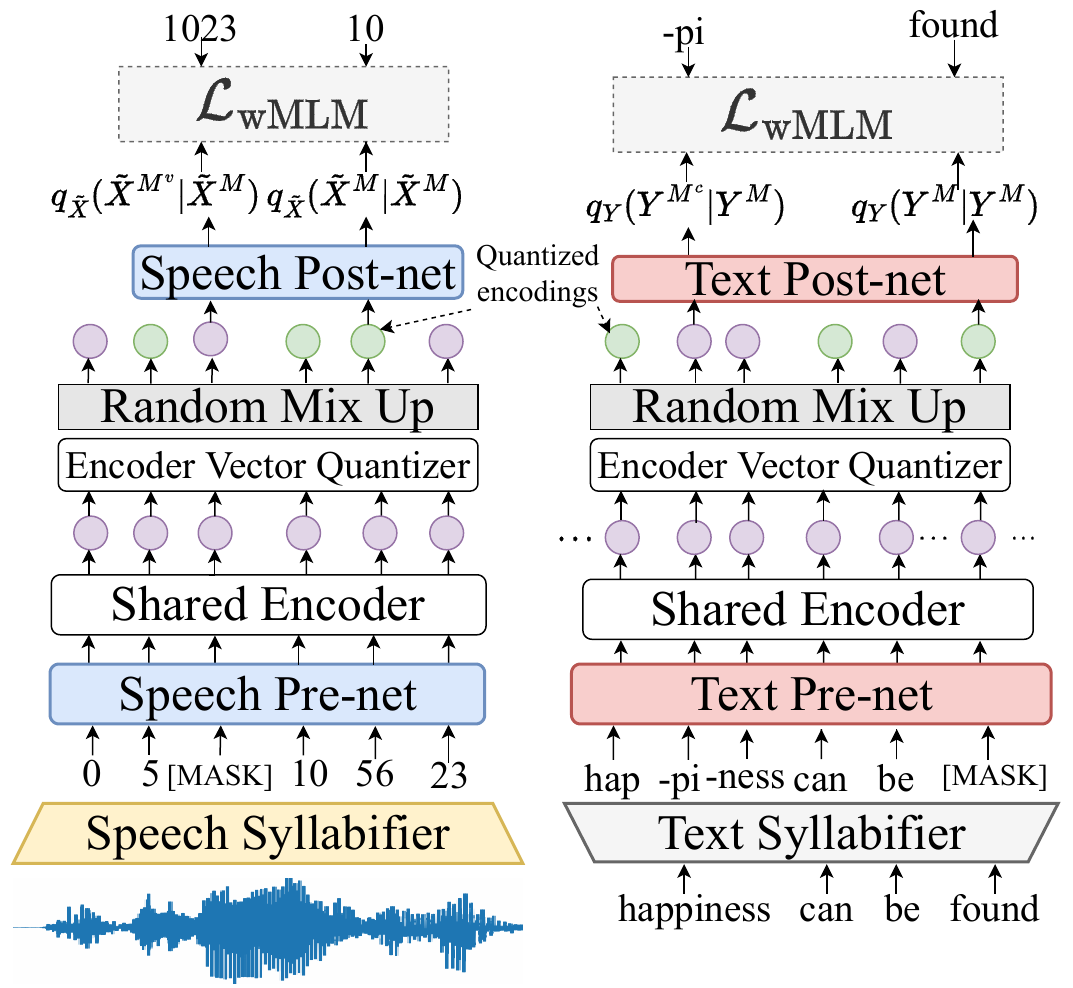}
        \caption{MLM-based stages}
    \end{subfigure}
    \begin{subfigure}{0.3\textwidth}
        \centering
        \includegraphics[width=0.9\textwidth]{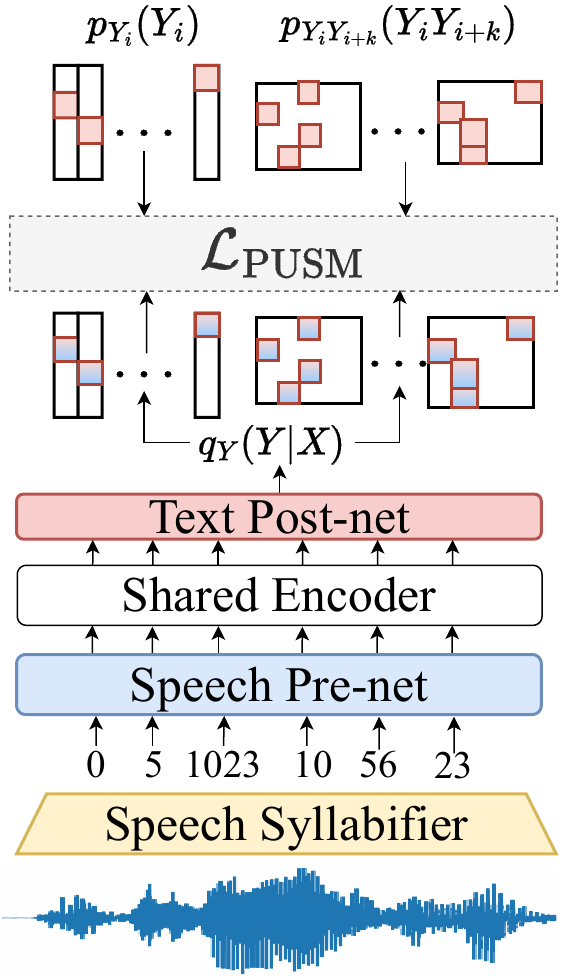}
        \caption{PUSM stage}
    \end{subfigure}
    \figvsp
    \caption{\textbf{Overall architecture of \sylcipher/}. Gray boxes are fixed during training. (a) MLM-based stages: learn a compressed joint semantic space with a shared encoder and random mix-up. (b) PUSM stage: align speech and text spaces by matching lower-order marginals of their distributions.} %(c) Speech syllabifier: refines syllable boundaries and tokens during the JE2E stage.}
    \figvsp
    \label{fig:sylcipher}
\end{figure}
% JSTTI
In this section, we describe \textbf{SylCipher}, our proposed model for syllable-level UASR. We first present its architecture and training objective, then justify the design theoretically, and finally introduce several practical modifications for training and inference.

\subsecvsp
\subsection{Training: UASR via information compression}
\begin{wrapfigure}{r}{0.24\textwidth}
    \centering
    \includegraphics[width=0.24\textwidth]{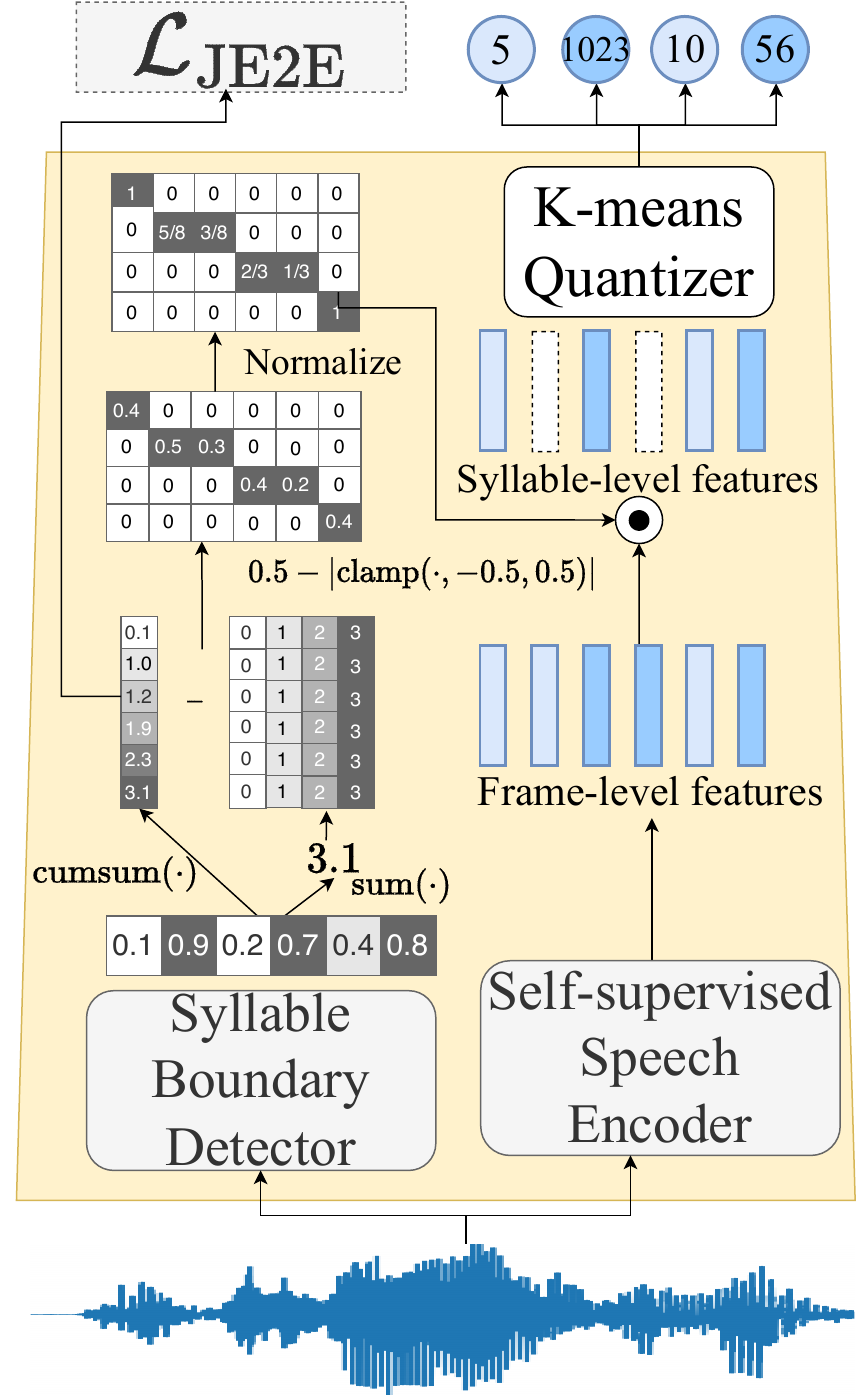}
    \figvsp
        \caption{Speech syllabifier}
        \label{fig:sylcipher_c}
    \figvsp
\end{wrapfigure}

\subsecvsp
As shown in Figure~\ref{fig:sylcipher}, SylCipher is an encoder-only language model with a \emph{shared encoder} for speech and text modalities. To project both modalities into a joint embedding space, we use two uni-modal \emph{\prenet/s}: $e_{\tilde{X}}:\gX^T\mapsto\sR^{L\times d}$ for speech and $e_Y:\gY^L\mapsto\sR^{L\times d}$ for text, each implemented as a linear embedding layer. Before the speech \prenet/, a \emph{speech syllabifier} converts the frame-level feature vectors into a syllable-level sequence. It consists of: (i) a \emph{differentiable soft-pooler} $m:\gX^T\mapsto\gX^L$ that aligns speech with text on the syllable level (ii) a tokenizer $c:\gX^L\mapsto \tilde{\gX}^L$ to discretizes speech into syllable-like units. Thus the speech \prenet/ is 
\begin{align}
    e_{\tilde{X}}(\tilde{X}) = e_{\tilde{X}}\circ c\circ m(X),
\end{align}
where $f\circ g(x):=f(g(x))$ for any functions $f,g$, and $\tilde{X}:= c\circ m(X)\in\tilde{\gX}^L.$ 
The soft-pooler first estimates the boundary probabilities $b(X)\in [0, 1]^T$ by learning from an unsupervised syllable detector  Sylber~\citep{Cho2025-sylber}, then constructs a pooling mask $a(X)\in [0, 1]^{L\times T}$, as illustrated in Figure~\ref{fig:sylcipher_c}:
\begin{equation}
\begin{aligned}\label{eq:softpooler}
    % a_{it}(X) := \epsilon-\left|\texttt{clamp}\brac{i-\sum_{\tau\leq t}b_{\tau}(X),-\epsilon,\epsilon}\right|, \quad m_i(X) := \sum_{t=1}^T \frac{a_{it}(X)}{\sum_{\tau=1}^T a_{i\tau}(X)}X_t, 
    \tilde{a}_{it}(X) := \sigma_{\epsilon}\brac{i-\sum_{\tau\leq t}b_{\tau}(X)},&\quad a_{it}(X)=\frac{\tilde{a}_{it}(X)}{\sum_{\tau=1}^T \tilde{a}_{i\tau}(X)},\quad m_i(X) := \sum_{t=1}^T a_{it}(X)X_t, 
\end{aligned}
\end{equation}
where $\sigma_{\epsilon}(x):=\epsilon - |\mathrm{clamp}(x,-\epsilon,\epsilon)|$ and $\mathrm{clamp}(x, a, b)=x$ for $x\in [a, b]$, $a$ if $x < a$ and $b$ otherwise. The hyperparameter $\epsilon$ controls the \emph{tapering speed} of pooling weights outside each syllabic segment. 
% Intuitively, the unnormalized pooling weights $a_i$'s keep track of whether a frame is inside segment $i$ by counting the number of segments up to time $t$, and add a heavy penalty to it when it is not. 
This creates sparse pooling weights that avoid overflow/underflow issues common in earlier soft-pooling methods with $\sigma_{\epsilon}(x)=\tanh(x/\epsilon)$~\citep{Bhati2022-scpc} and $\sigma_{\epsilon}(x)=\exp(x/\epsilon)$~\citep{wang2024unsupervised}.
% Similar operations have been used before with $\sigma_{\epsilon}(x)=\tanh(x/\epsilon)$~\citep{Bhati2022-scpc} and $\sigma_{\epsilon}(x)=\exp(x/\epsilon)$~\citep{wang2024unsupervised}.
% but often involve nested sigmoids or softmax operations, leading to floating point overflow/underflow issues and training instability. Instead, our approach create a \emph{sparse} pooling matrix, thus avoiding extreme logit values. 
To contextualize embeddings, both modalities are processed by a shared encoder $f:\sR^{T\times d}\mapsto\sR^{T\times d}$:
% them to a joint embedding space, a shared encoder $f:\sR^{T\times d}\mapsto\sR^{T\times d}$ is used:
\begin{align}
    f_{\tilde{X}}(\tilde{X}) := f\circ e_{\tilde{X}}(\tilde{X}),\quad f_Y(Y) := f\circ e_Y(Y).
\end{align}
In practice, a differentiable \km/ vector quantizer~\citep{Ni2025-jstti} is used for the tokenizer $c$ and a multi-layer transformer~\citep{Vaswani2017} is used for the shared encoder. 

\paragraph{Distribution matching.} % During training, no speech-text pairs are available, and thus the best we can do to capture the information of the speech and text modalities is to estimate their unimodal \pdf/s, $p_{\tilde{X}}$ and $p_Y$. 
Without paired speech-text data, we approximate each unimodal distribution $p_{\tilde{X}}$ and $p_Y$. Let $g_{\tilde{X}},\, g_Y:\sR^{d\times L}\mapsto[0,1]$ be speech and text \emph{\postnet/s} such that
\begin{align}\label{eq:postnet}
    q_{Z}(Z) &:= g_Z\circ f_Z(Z):=\prod_{i=1}^L g_{Z,i}(Z_i|f_Z(Z^{\{1,\cdots,i-1\}}))=\prod_{i=1}^{L}\frac{e^{W^Z_{iZ_i}f_Z(Z^{\{1,\cdots,i-1\}})}}{\sum_{z}e^{W^Z_{iz}f_Z(Z^{\{1,\cdots,i-1\})}}},
\end{align}
where $Z=\tilde{X}$ or $Y$ with equal probabilities is a \emph{mixture random sequence}, and $Z^M_i:=Z_i,\,i\in M$ and \texttt{[MASK]} otherwise for some index set $M$, and $W^Z_i\in \sR^{\gZ\times dL}.$

Training minimizes \emph{Kullback–Leibler (KL) divergence} to unimodal distributions:
\begin{align}
    \min_{q_{\tilde{X}},q_Y}\KL(p_{\tilde{X}}||q_{\tilde{X}})+\KL(p_Y||q_Y)=\KL(p_{\tilde{X}}||g_{\tilde{X}} \circ f_{\tilde{X}})+\KL(p_Y||g_Y\circ f_Y).
\end{align}
% However, one issue immediately follows: nothing can prevent the model to learn \emph{disjoint} regions of the joint embedding space for speech and text, even if it estimate the unimodal \pdf/s perfectly, due to the lack of limit imposed on the capacity of the model. To fix this issue, we limit the capacity of the model via the following constrained optimization problem:
To prevent speech and text from occupying disjoint embedding regions, we further constrain the shared encoder \emph{entropy}. Let  $H(X)$ denotes the entropy of discrete random variable $X$, consider
\begin{align}\label{eq:regularized_distribution_matching}
    &\min_{f_{\tilde{X}}, g_{\tilde{X}}, f_Y, g_Y}\KL(p_{\tilde{X}}||g_{\tilde{X}}\circ f_{\tilde{X}})+\KL(p_Y||g_Y\circ f_Y)\quad\text{s.t.}\quad H(f_{Z}(Z))\leq H(Y),
\end{align} 

\paragraph{Theoretical guarantee.} We prove  that under regularity conditions, \eqref{eq:regularized_distribution_matching} matches true and generated text distributions in the same way as GANs, but without unstable straight-through gradients. Thus zero-error UASR is achievable under conditions in \citep{wang-etal-2023-unsupasr-theory}.
\begin{theorem}\label{thm:main}
    Suppose $(f_{\tilde{X}}^*, g_{\tilde{X}}^*, f_Y^*, g_Y^*)$ minimize \eqref{eq:regularized_distribution_matching}, then under certain assumptions (See Appendix~\ref{app:proof_of_main}), $f_{\tX}^*$ and $f_Y^*$ are invertible and $q_{Y|X}^*(y|x):=\mathbbm{1}[f_Y^{*-1}\circ f_{\tilde{X}}^*\circ c \circ m(x)=y]$
    satisfies 
    \[
    \KL(p_Y||\E_X[q_{Y|X}^*])=0,\quad y^*(x)=\argmax_{y\in \gY^L} q^*_{Y|X}(y|x),\quad\forall x\in\gX^T.
    \]
\end{theorem}

\paragraph{Practical training.} In practice, \sylcipher/ implements unimodal masked language modeling (MLM) to approximate unimodal \pdf/s. Given a mask distribution $p_M$, we minimize the weighted loss:
\begin{align}\label{eq:mlm}
    \Ls_{\mathrm{wMLM}}(\theta_X,\theta_Y):=-\E_{Z\sim \frac{1}{2}p_{\tilde{X}}+\frac{1}{2}p_Y,\,M\sim p_M}\sbrac{\ln q_Z(Z^M|Z^{M^c})+\lambda \ln q_Z(Z^M|Z^M)},
\end{align}
where $\theta_X$ and $\theta_Y$ are speech-related and text-related parameters. Notice that $g_Y$ serves a dual role: it acts both as the text \postnet/ for unimodal MLM and as a \emph{cross-modal decoder}, defined as (similar for $q_X$) $q_Y(Y^M|Z^M):=\prod_{i\in M} g_{Y,i}(Y_i|f_Z(Z_i)).$ The additional term weighted by $\lambda$ balances these two functions, enabling $g_Y$ (or $g_X$) to reconstruct text (or speech) tokens from either masked text or speech inputs. To further constrain the entropy of the encoding outputs $f_Z(Z)$, we limit the transformer depth to two layers and apply \emph{random mix-up}~\citep{Ao2022-speecht5}. During training, a random subset of encoder representations is quantized to a finite set of code vectors before passed to the \postnet/s. This restricts the variability of $f_Z(Z)$ without significantly reducing its predictive capacity for masked tokens.

% JE2E stage
After pretraining with unsupervised syllable boundary labels $[\hat{b}_1(X),\cdots,\hat{b}_T(X)]$, we perform \emph{joint end-to-end} (JE2E) training. In this stage, the soft-pooler is trained jointly with the rest of the model, guided by an additional soft constraint on the predicted \emph{syllable counts}:
\begin{align}
    \Ls_{\mathrm{JE2E}}(\theta_X) := \E_{X}\left|\sum_{t=1}^T b_t(X)-\hat{b}_t(X)\right|,
\end{align}
which discourages both over- and under-segmentation.

% PUSM stage
While the MLM objective encourages \emph{implicit} distribution matching, we found that \emph{explicit} distribution matching further improves a saturated MLM system. To this end, we adopt positional unigram and skipgram matching (PUSM)~\citep{wang2024unsupervised,Ni2025-jstti}:
\begin{align}\label{eq:pusm}
    \Ls_{\mathrm{PUSM}}(\theta_X,\theta_Y)=\|\E_X\sbrac{q_{Y_i}(X)}- p_{Y_i}]\|_1+\sum_{k=1}^K\left\|\E_X\sbrac{q_{Y_iY_{i+k}}(X)}-p_{Y_iY_{i+k}}\right\|_1,
\end{align}
where $q_{Y}(x):=q_{Y}(\cdot|c\circ m(x))$ and $K$ is the maximal skip length. The first term matches unigram distributions at each position, while the second aligns skipgram distributions. To stabilize training, we disable random mix-up and approximate the empirical \pdf/s using the entire dataset as a single batch, reducing memory cost via gradient accumulation~\citep{Ni2025-jstti}. The overall \sylcipher/ objective combines all stages:
\begin{align}\label{eq:sylcipher}
    \Ls_{\mathrm{SylCipher}} := \lambda_{\mathrm{wMLM}}\Ls_{\mathrm{wMLM}}+\lambda_{\mathrm{JE2E}}\Ls_{\mathrm{JE2E}}+\lambda_{\mathrm{PUSM}}\Ls_{\mathrm{PUSM}}.
\end{align}
Because the PUSM stage requires different batch sizes, we adopt an \emph{iterative training schedule} rather than fully end-to-end optimization. Specifically:
\begin{enumerate}[leftmargin=2em, labelsep=0.5em, itemindent=0.0em]
    \item \textbf{Fixed boundary stage}: freeze the boundary detector $b$ and set $\lambda_{\mathrm{JE2E}}=\lambda_{\mathrm{PUSM}}=0$;
    \item \textbf{JE2E stage}: enable $\lambda_{\mathrm{JE2E}}>0$ to refine segmentation;
    \item \textbf{PUSM stage}: disable $\lambda_{\mathrm{wMLM}}$, enable $\lambda_{\mathrm{PUSM}}>0$ for explicit distribution matching. 
\end{enumerate}

% Inference stage
\subsecvsp
\subsection{inference}
\subsecvsp
During inference, the ASR system cascades the speech syllabifier, speech pre-net and the text post-net as done in the PUSM stage:
\begin{align}
    y(X):=\argmax_{y\in\gY^L} q_{Y}(y|X).
\end{align}
We observed that using the \emph{first} transformer layer of the shared encoder, instead of the last, improves performance, consistent with findings at the word-level~\citep{Ni2025-jstti}. This may be due to over-contextualization in later layers. Finally, replacing each \texttt{<OOV>} token with the second most likely prediction reduces CER by about $1\%$ compared to simply discarding \texttt{<OOV>}s.

\secvsp
\section{Experiments}\label{sec:exp}
\secvsp
%\begin{table}[ht]
%    \caption{\textbf{Syllable boundary detection results on \aishell/}. The superscript for F1 scores is tolerance threshold in ms and the tolerance is 50ms for other metrics. P., Re. and R stand for precision, recall and R-value respectively. Models labeled with ``ft.'' are further finetuned on \aishell/. The  SylCipher boundary detector is trained for the \emph{without-tone} setting.}
%    \label{tab:syl_bnd_aishell}
%    \centering
%    \begin{tabular}{l|ccccc}
%    \toprule
%         & \textbf{F1$^{50}$ ($\uparrow$)} & \textbf{F1$^{20}$ ($\uparrow$)} & \textbf{P. ($\uparrow$)} & \textbf{Re. ($\uparrow$)} & \textbf{R ($\uparrow$)} \\
%    \midrule
%    \midrule
%    SylBoost~\citep{Baade2025-syllablelm} & & & \\
%    Sylber~\citep{Cho2025-sylber} & 64.2 & 35.9 & 63.3 & 65.2 & 69.1 \\
%    Sylber ft. & 73.8 & 49.8 & 74.1 & 73.5 & 77.7 \\
%    \midrule
%    SylCipher (Ours, Sylber ft.+JE2E) &  \\
%    \bottomrule
%    \end{tabular}
%\end{table}
Section~\ref{sec:dataset} introduces the datasets used for our experiments, followed by the syllabification steps for speech and text preprocessing in Section~\ref{sec:preproc}. Section~\ref{sec:uasr_result} presents the main UASR results, and Section \ref{sec:bd_result} discusses \sylcipher/'s boundary refinement ability. Section~\ref{sec:ablation} provides ablation studies on key design choices. Additional implementation details of our method and the baselines are given in Appendix~\ref{app:implementation}.

\subsecvsp
\subsection{Datasets}\label{sec:dataset}
\subsecvsp
We evaluate \sylcipher/ on three datasets. First, we train on the 460-hour clean subset of LibriSpeech~\citep{Panayotov15-LibriSpeech}, a standard UASR benchmark of audiobook recordings. 
Second, to test domain generalization, we train another model on SpokenCOCO~\citep{hsu2021textfree}, which contains 742 hours of spoken image captions. Third, to study languages with syllabic structures significantly different from English, we apply \sylcipher/ to Mandarin using \aishell/~\citep{shi21c_interspeech}, which has 85 hours of read speech. We follow the same LibriSpeech split as in \citep{Ni2025-jstti}, and use the standard splits for SpokenCOCO and \aishell/. For LibriSpeech and SpokenCOCO, we consider both the \emph{matched} setting, where empirical speech and text \pdf/s can be matched exactly, and the more realistic \emph{unmatched} setting where they cannot. 
In the matched case, we use paired speech-text datasets with pairings removed; in the unmatched case, LibriSpeech uses  LibriLM~\citep{panayotov2015librispeech} with overlapping text removed, while SpokenCOCO is randomly split in half, with one half treated as  speech-only and the other half treated as text-only. For \aishell/, we consider only the matched setting and compare models trained with/without tone labels. For all datasets, we apply a voice activity detector\footnote{https://github.com/wiseman/py-webrtcvad.git} to improve alignment.

\subsecvsp
\subsection{Speech and text syllabification}\label{sec:preproc}
\subsecvsp
For English text, we use \pyphen/\footnote{https://github.com/Kozea/Pyphen}, a rule-based hyphenation tool re-purposed for syllabification without a G2P. If \pyphen/ produces only a single chunk for a long word, we apply a simple rule-based fallback (Appendix~\ref{app:code_pyphen+}). We denote this combined approach as \pyphen/+. We also experiment with other G2P-free approaches such as byte-pair encoding (BPE)~\cite{liu-etal-2025-superbpe,sennrich-etal-2016-neural}, and find \sylcipher/ robust to syllabification errors. To avoid long-tail distributions, we keep only the top-2048 most frequent English syllables and replace the rest with a special \texttt{<OOV>} token (replacing $7\%$ of tokens in LibriSpeech and $2\%$ in SpokenCOCO. For Mandarin, we use the Pinyin of each Chinese character as a syllable, with or without tone labels. We keep the top-1024 most frequent syllables, covering 99.5\% of occurrences. For speech syllabification, we use \km/ clustering on syllable-level speech features created by mean pooling within the Sylber boundaries, with codebook size equal to the number of non-\texttt{<OOV>} text tokens.
\begin{table}[t]
    \caption{\textbf{UASR results on LibriSpeech (clean subsets) and SpokenCOCO}. Inside the bracket lists the unsupervised boundary used for each model. \emph{Student} stands for the student model used during the self-training stage using pseudo-labels from each model. For tokens used for the text data, \emph{Char.} stands for characters and \emph{Syllable} stands for syllable-level tokens converted using the Pyphen+ syllabifier without using a G2P. CER stands for character error rate. * indicates evaluation on LibriSpeech dev-clean instead.}
    \figvsp
    \begin{subtable}{1.0\textwidth}
    \centering
    \caption{UASR results on LibriSpeech (clean subsets)}
    \figvsp
    \resizebox{0.9\textwidth}{!}{
    \begin{tabular}{llcccc}
    \toprule
        \multirow{2}{*}{\textbf{Model}} &  \multirow{2}{*}{\textbf{Student}} & \multirow{2}{*}{\textbf{Token}} & \multicolumn{1}{c}{\textbf{Matched}} & \multicolumn{1}{c}{\textbf{Unmatched}} \\
        & & & \textbf{CER ($\downarrow$)} %& \textbf{WER ($\downarrow$)}
        & \textbf{CER ($\downarrow$)} %& \textbf{WER ($\downarrow$)}
        \\
    \midrule
    \midrule   
    \multicolumn{5}{c}{\emph{G2P-based approach}} \\
    \midrule
    \multirow{1}{*}{\wtovu/*~\citep{Baevski2021-wav2vec-u}} & None & Phone & - & 13.3\\
    \multirow{1}{*}{\wtovu/ 2.0*~\citep{Liu2023-wav2vecu2}} & None & Phone & - & 12.2\\
    \multirow{1}{*}{\reborn/*~\citep{Tseng2024-reborn}} & None & Phone & - & 8.3 \\
    \midrule
    \multicolumn{5}{c}{\emph{G2P-free approach}} \\
    \midrule
        \multirow{2}{*}{\wtovu/~\citep{Baevski2021-wav2vec-u}}  & None & Char. & 35.6 & 43.3 \\
        & \wtov/ & Char. & 33.8 & 42.1 \\
        \multirow{1}{*}{\reborn/~\citep{Tseng2024-reborn}} & None & Char. & 37.8 & 76.6\\
        JSTTI (forced align) & None & Char & 81.4 & 81.1\\
        \multirow{1}{*}{JSTTI~\citep{Ni2025-jstti}} & None & Word & 49.5 & 54.2\\
        \multirow{2}{*}{PUSM (Sylber)~\citep{wang-etal-2023-unsup-speech2sign}} & None & Syllable & 35.5 & 57.7 \\
        & \wtov/ & Syllable & 33.0 & 54.7\\
        \midrule
        \sylcipher/ (Ours, forced align) & None & Syllable & 38.5 & 46.4 \\
        
        \sylcipher/ (Ours, Sylber) & None & Syllable & 43.5 & 48.6\\
        \sylcipher/ (Ours, Sylber+JE2E) & None & Syllable & 39.2 & 46.8 \\
        \multirow{2}{*}{\sylcipher/ (Ours, Sylber+JE2E+PUSM)} & None & Syllable & \textbf{21.8} & \textbf{35.9} & \\
        & \wtov/ & Syllable & \textbf{17.5} & \textbf{33.3}\\
    \bottomrule
    \end{tabular}}
    \label{tab:uasr_libri}
    \end{subtable}
    \begin{subtable}{1.0\textwidth}
    \centering
    \caption{UASR results on SpokenCOCO}
    \figvsp
    \resizebox{0.9\textwidth}{!}{
    \begin{tabular}{llcccc}
    \toprule
        \multirow{2}{*}{\textbf{Model}} &  \multirow{2}{*}{\textbf{Student}} & \multirow{2}{*}{\textbf{Token}} & \multicolumn{1}{c}{\textbf{Matched}} & \multicolumn{1}{c}{\textbf{Unmatched}} \\
        & & & \textbf{CER ($\downarrow$)} %& \textbf{WER ($\downarrow$)}
        & \textbf{CER ($\downarrow$)} %& \textbf{WER ($\downarrow$)}
        \\
    \midrule
    \midrule
        \multirow{2}{*}{\wtovu/~\citep{Baevski2021-wav2vec-u}}  & None & Char. & 45.0 & 45.2 \\
        & \wtov/ & Char. & 46.3 & 35.3\\
        %\multirow{2}{*}{\reborn/~\citep{Tseng2024-reborn}} & None & Char. &  & \\
        % & \wtov/ & Char. \\
        
        JSTTI & None & Char. & 78.3 & 100 \\
        \multirow{1}{*}{JSTTI~\citep{Ni2025-jstti}} & None & Word & 64.5 & 64.5\\
        \multirow{2}{*}{PUSM (Sylber)~\citep{wang-etal-2023-unsup-speech2sign}} & None & Syllable & 41.5 & 41.3 \\
        & \wtov/ & Syllable & 34.7 & 34.3 \\
        \midrule
        \sylcipher/ (Ours, Sylber) & None & Syllable & 34.9 & 36.1 \\
        \sylcipher/ (Ours, Sylber+JE2E) & None & Syllable & 31.2 & 32.4 \\
        \multirow{2}{*}{\sylcipher/ (Ours, Sylber+JE2E+PUSM)} & None & Syllable & \textbf{23.4} & \textbf{26.8} \\
        & \wtov/ & Syllable & \textbf{13.9} & \textbf{17.6}\\
    \bottomrule
    \end{tabular}}
    \label{tab:uasr_scoco}
    \end{subtable}
    \figvsp
\end{table}

\begin{table}[t]
    \centering
    \caption{\textbf{UASR results on AISHELL-3 test set}. Inside the bracket lists the unsupervised boundary used for each model. ``Student'' stands for the student model used during the self-training stage using pseudo-labels from each model. \emph{Init./Final} refers to initials and finals in the Chinese phonetic alphabet. For text data tokens, \emph{Syllable} refers to the pinyin representation of each Chinese character, while \emph{Phone} denotes the individual letters within the pinyin tokens. PER stands for phone error rate.}
    \figvsp
    \resizebox{0.9\textwidth}{!}{
    \begin{tabular}{llcccc}
    \toprule
        \multirow{2}{*}{\textbf{Model}} &  \multirow{2}{*}{\textbf{Student}} & \multirow{2}{*}{\textbf{Token}} & \multicolumn{1}{c}{\textbf{w/o Tone}} & \multicolumn{1}{c}{\textbf{w. Tone}} \\
        & & & \textbf{PER ($\downarrow$)} %& \textbf{WER ($\downarrow$)}
        & \textbf{PER ($\downarrow$)} %& \textbf{WER ($\downarrow$)}
        \\
    \midrule
    \midrule
        \multirow{1}{*}{\wtovu/~\citep{Baevski2021-wav2vec-u}}  & None & Phone & 74.9 & 76.2 \\
        % & None & Init./Final & \liming{run by Kai-Wei} & \liming{run by Kai-Wei}\\
        %\multirow{2}{*}{\reborn/~\citep{Tseng2024-reborn}} & None & Phone &  & \\
        % & \wtov/ & Phone \\
        
        JSTTI & None & Init./Final & 96.4 & 100\\
        \multirow{1}{*}{JSTTI~\citep{Ni2025-jstti}} & None & Word & 83.2 & 169\\
        \multirow{2}{*}{PUSM (Sylber)~\citep{wang-etal-2023-unsup-speech2sign}} & None & Syllable & 28.4 & 26.5 \\
        & \wtov/ & Syllable & 18.5 & 14.9\\
        \midrule
        \sylcipher/ (Ours, forced align) & None & Syllable & 38.1 & 38.9\\
        
        \sylcipher/ (Ours, Sylber) & None & Syllable & 44.6 & 48.3 \\
        \sylcipher/ (Ours, Sylber+JE2E) & None & Syllable & 41.7 & 45.1 \\
        \multirow{2}{*}{\sylcipher/ (Ours, Sylber+JE2E+PUSM)} & None & Syllable & \textbf{26.9} & \textbf{24.9} \\
        & \wtov/ & Syllable & \textbf{15.3} & \textbf{12.2} \\
    \bottomrule
    \end{tabular}}
    \label{tab:uasr_aishell3}
    \figvsp
\end{table}
\subsecvsp
\subsection{Results: UASR}\label{sec:uasr_result}
\subsecvsp
We compare \sylcipher/ with UASR systems that differ in token type, training objective, and architecture. 

\textbf{Word-level:} \emph{JSTTI}~\citep{Ni2025-jstti}, the state-of-art word-level UASR system, which is GAN-free and architecturally similar to ours, using 2048 non-\texttt{<OOV>} words.

\textbf{Character-level:} \emph{\wtovu/}~\citep{Baevski2021-wav2vec-u}, a strong GAN-based system; \emph{\reborn/}~\citep{Tseng2024-reborn}, a state-of-the-art phoneme-based GAN system; and a \emph{phone-level JSTTI}, trained with phoneme boundaries and 128 speech clusters (comparable to the number of character types).

\textbf{Syllable-level:} \emph{PUSM}~\citep{wang-etal-2023-unsup-speech2sign}, adapted from word-level UASR to syllables using the same syllable boundary detector and syllabifier as \sylcipher/. 

We also report results with \emph{self-training}~\cite{Chen2019-uasr,Baevski2021-wav2vec-u}, where a \wtov/~\cite{Baevski2020-wav2vec2} student is further finetuned by distilling character (grapheme)-level pseudo-labels from a UASR system.  
Performance is measured by character error rate (CER) for English and phone error rate (PER) for Mandarin, as both are tokenization-independent and easily comparable. 

% Which token type is better
\noindent\textbf{Syllable-level modeling performs best under G2P-free setting.} Table~\ref{tab:uasr_libri} summarizes results on LibriSpeech. Among baselines, PUSM performs best in the matched setting, while \wtovu/ is strongest in the unmatched setting, though its performance is limited by the lack of G2P. Unlike phonemized training, \reborn/ trained directly on raw characters performs worse than \wtovu/, suggesting sensitivity to misalignment between speech and text. By contrast, \sylcipher/ with all three stages (Sylber+JE2E+PUSM) achieves 21.8\% CER (matched) and 35.9\% (unmatched), outperforming all baselines. Compared to the best-in-average system \wtovu/ (35.6\%/43.3\%), \sylcipher/ reduces by CER by 40\% (matched) and 17\% (unmatched) relative. 
Both syllable-level models (PUSM and \sylcipher/) outperform word- or character-level models, confirming that speech-text alignment is the best at the syllable level. Word-level JSTTI performs worst due to poor rare-word coverage (\texttt{<OOV>}s $\approx$ 17\%), and adapting JSTTI to phone-level degrades further, likely because phone clusters are noisier than syllable or word clusters.

% effect of iterative training
\noindent\textbf{Iterative training helps.} Stage-wise training shows progressive improvements: JE2E reduces CER modestly, while PUSM yields the largest gains (44\% and 23\% relative reductions over JE2E in matched/unmatched settings). Combining MLM-based stages with PUSM outperforms PUSM-only training by 33-34\% relative, as MLM provides necessary initialization for PUSM to converge. Indeed, PUSM alone fails when \sylcipher/ is randomly initialized, likely due to transformer training instability. Lastly, using unsupervised Sylber boundaries performs nearly as well as forced alignment, suggesting robustness to segmentation noise.
% Effect of self-training
Self-training consistently improves performance, especially for stronger models. For \sylcipher/, CER is further reduced by 20\% relative. 

% Repeat for SCOCO
\noindent\textbf{Syllables are robust to domain shifts.} Table~\ref{tab:uasr_scoco} reports results on SpokenCOCO.  \sylcipher/ again outperforms all baselines, surpassing PUSM by 32\% and \wtovu/ by 49\% relative CER after all stages. The margin is larger than LibriSpeech, especially in the unmatched setting. Notably, even after the first training stages, \sylcipher/ already outperforms \wtovu/  by 22\%. While \wtovu/ suffers from domain mismatch at the character level, both syllable-level methods perform better, suggesting syllable units are more robust to domain shifts. Self-training further improves \sylcipher/ by 39\% relative CER, likely due to higher syllable coverage reducing insertion/deletion errors. Moreover, performance degrades only slightly in the unmatched setting, indicating the sharper degradation on LibriSpeech stems from domain mismatch between its speech and text corpora.

\noindent\textbf{Syllable-level UASR works for Mandarin.} Table~\ref{tab:uasr_aishell3} presents results on Mandarin. Here, syllable-level models converge even without boundary refinement, while phone- and word-level approaches struggle, confirming that syllables are the most natural alignment unit for Mandarin. Compared to the best baseline (PUSM), \sylcipher/ achieves over 5.5\% relative PER reduction before self-training and 17\% after. Interestingly, including tone labels does not harm performance--in fact, PER improves after the PUSM stage--suggesting that once phoneme labels are predicted correctly, tone prediction is also reliable.

\begin{table}[ht]
    \caption{\textbf{Syllable boundary detection results on LibriSpeech and SpokenCOCO}. The superscript for F1 scores is tolerance threshold in ms and the tolerance is 50ms for other metrics. P., Re. and R stand for precision, recall and R-value respectively. \sylcipher/ is trained under unmatched settings.}
    \centering
    \begin{subtable}{1.0\textwidth}
    \centering
    \figvsp
    \caption{Boundary detection results on LibriSpeech}
    \figvsp
    \begin{tabular}{l|ccccc}
    \toprule
         & \textbf{F1$^{50}$ ($\uparrow$)} & \textbf{F1$^{20}$ ($\uparrow$)} & \textbf{P. ($\uparrow$)} & \textbf{Re. ($\uparrow$)} & \textbf{R ($\uparrow$)} \\
    \midrule
    \midrule
    Feat-Sim~\citep{Peng2023-syllable}  & 47.3 & 24.7 & 46.6 & 48.0 & 54.4 \\
    SDHuBERT~\citep{cho2023-sdhubert} & 66.1 & 32.2 & 64.9 & 67.4 & 70.7 \\
    SylBoost~\citep{Baade2025-syllablelm} & 73.2 & \underline{44.6} & 72.1 & 74.4 & 76.9\\
    Sylber~\citep{Cho2025-sylber} & \underline{83.4} & 44.1 & \underline{84.8} & \underline{84.1} & \underline{86.4}\\
    \midrule
    SylCipher (Ours, Sylber+JE2E) & \textbf{86.1} & \textbf{50.8} &  \textbf{86.6} & \textbf{86.1} & \textbf{88.1} \\
    \bottomrule
    \end{tabular}
    \label{tab:syl_bnd_libri}
    \end{subtable}
    
    \begin{subtable}{1.0\textwidth}
    \centering
    \caption{Boundary detection results on SpokenCOCO}
    \figvsp
    \begin{tabular}{l|ccccc}
    \toprule
         & \textbf{F1$^{50}$ ($\uparrow$)} & \textbf{F1$^{20}$ ($\uparrow$)} & \textbf{P.} ($\uparrow$) & \textbf{Re.} ($\uparrow$) & \textbf{R ($\uparrow$)} \\
    \midrule
    \midrule
    Feat-Sim~\citep{Peng2023-syllable}  & \underline{60.3} & - & 57.4 & \underline{63.6} & \underline{64.3}\\
    SylBoost~\citep{Baade2025-syllablelm} & 55.6 & \underline{30.2} & 48 &  \textbf{65.2} & 50.8 \\
    Sylber~\citep{Cho2025-sylber} & 53.5 & 28.6 & 52.3 & 54.9 & 59.6 \\
    \midrule
    SylCipher (Ours, Sylber+JE2E) & \textbf{62.3} & \textbf{31.0} & \textbf{61.1} & \underline{63.6} & \textbf{67.4}	\\
    \bottomrule
    \end{tabular}
    \label{tab:syl_bnd_scoco}
    \end{subtable}
    \figvsp
    \figvsp
\end{table}

\subsecvsp
\subsection{Results: Unsupervised syllable boundary detection}\label{sec:bd_result}
\subsecvsp
To better understand the JE2E stage, we compare the syllable boundary detection performance against the teacher model Sylber~\citep{Cho2025-sylber} (which provides \sylcipher/'s initial boundaries) and other unsupervised approaches including Feat-Sim~\citep{Peng2023-syllable}, SDHuBERT~\citep{cho2023-sdhubert}, SylBoost~\citep{Baade2025-syllablelm}. On LibriSpeech, Sylber is the strongest baseline on most metrics, except for the 20ms F1 score where SylBoost performs best. On SpokenCOCO, SylBoost outperforms Sylber on three of five metrics. However, our preliminary analysis show that SylBoost often over-segments, producing too many syllables and causing cross-modal misalignment and training instability in \sylcipher/. In contrast, Sylber predicts syllable counts closer to ground truth, making it more suitable as an initialization for UASR. When refined through JE2E, \sylcipher/ improves upon its Sylber teacher, achieving +14\% relative F1 (20ms tolerance) and +3\% relative F1 (50ms) on LibriSpeech. On SpokenCOCO, it surpasses Sylber by +15\% F1 and +11\% R-value, and outperforms the best baseline Feat-Sim by +3\% relative F1 score and +4.6\% R-value. These results suggest that unpaired text provides additional guidance on syllable boundary detection. Visualizations of the speech-text alignment predicted by \sylcipher/ can be found in Appendix~\ref{app:spec_libri}.  
\begin{figure}[t]
\centering
\begin{subfigure}{0.6\textwidth}
    \centering
    \caption{Effect of syllabifier type}
    \resizebox{0.8\textwidth}{!}{
    \begin{tabular}{lccc}
         \toprule
         \textbf{Model} & \multirow{1}{*}{\textbf{Tokenizer}} & \multirow{1}{*}{\textbf{Resource}} & \textbf{SER ($\downarrow$)} \\
         \midrule
         \midrule
         \sylcipher/ (Sylber) & \pyphen/+ & Medium & \underline{48.9}\\
         \sylcipher/ (Sylber+JE2E) & \pyphen/+ & Medium & \textbf{44.9}\\
         \sylcipher/ (Sylber) & Syllabify & High & 51.3\\
         \sylcipher/ (Sylber) & BPE+ & Low & 52.5\\
         \sylcipher/ (Sylber+JE2E) & BPE+ & Low & 49.9\\
         %Clamp & LibriSpeech & Yes & 1024 & \pyphen/+ & \liming{run by Saurabh; make a figure} \\
         %Clamp & LibriSpeech & Yes & 1280 & \pyphen/+ & \liming{run by Saurabh; make a figure} \\
         %Clamp & LibriSpeech & Yes & 1536 & \pyphen/+ & \liming{run by Saurabh; make a figure} \\
         %Clamp & LibriSpeech & Yes & 1792 & \pyphen/+ & \liming{run by Saurabh; make a figure} \\
         %Clamp & SpokenCOCO & Yes & 2048 & \pyphen/+ & \textbf{32.6}\\
         %Tanh & SpokenCOCO & Yes & 2048 & \pyphen/+ & 50.0\\
         %Softmax & SpokenCOCO & Yes & 2048 & \pyphen/+ & 58.4 \\
         %Clamp & SpokenCOCO & No & 2048 & \pyphen/+ & \textbf{33.4}\\
         %Tanh & SpokenCOCO & No & 2048 & \pyphen/+ & 52.0 \\
         %Softmax & SpokenCOCO & No & 2048 & \pyphen/+ & 62.4\\
         %Clamp & SpokenCOCO & Yes & 2048 & Syllabify & 37.1 \\
         %Clamp & AISHELL-3 & Yes & 431 & Pinyin (w/o tone) & \textbf{52.5}\\
         %Tanh & AISHELL-3 & Yes & 431 & Pinyin (w/o tone) & 66.5\\
         %Softmax & AISHELL-3 & Yes & 431 & Pinyin (w/o tone) & 56.1\\
         \bottomrule
    \end{tabular}}
    \label{tab:ablation}    
    \end{subfigure}
    \begin{subfigure}{0.4\textwidth}
        \centering
        \includegraphics[width=0.99\textwidth]{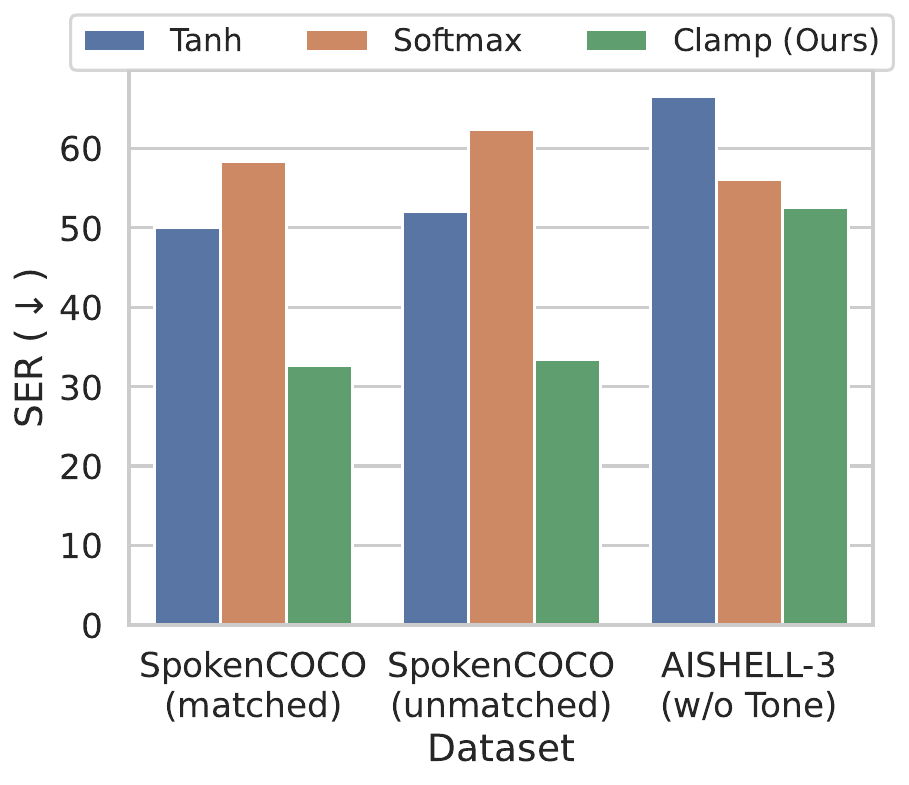}
        \figvsp
        \caption{SER vs. pooler type across datasets}
        \figvsp
        \label{fig:ser_vs_pooler}
    \end{subfigure}
    \begin{subfigure}{0.4\textwidth}
        \centering
        \includegraphics[width=0.99\textwidth]{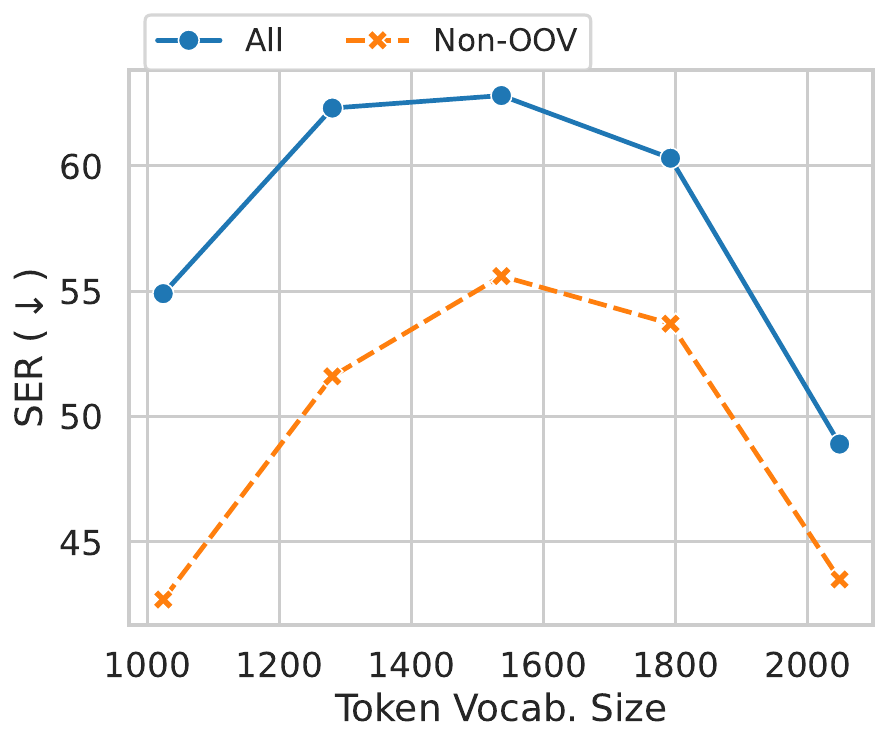}
        \figvsp
        \caption{SER vs. token vocabulary size\liming{running}}
        \figvsp
        \label{fig:ser_vs_token_vocab}
    \end{subfigure}
    \caption{\textbf{Ablation studies on the effect of syllabifier type, pooling type and token vocabulary size on \sylcipher/ UASR performance}. (a) The effect of different pooler named after the type of $\sigma_{\epsilon}$ function used in \eqref{eq:softpooler}.(b) The effect of the vocabulary size of non-\texttt{<OOV>}-tokens on \sylcipher/ performance during the fixed-boundary stage on LibriSpeech (matched). (c) Effect of syllabifiers with difference resource requirements on \sylcipher/ performance during the fixed-boundary stage on LibriSpeech (matched).}
    \label{fig:ablation}
    \figvsp
\end{figure}
\subsecvsp
\subsection{Ablation studies}\label{sec:ablation}
\subsecvsp
%\liming{An ablation table showing the effect of tokenization method, vocab size, encoder layer used, diff bnd mechanism used and a figure showing OOV decoding methods; with/without mixup}
We conduct ablation studies on various components and design choices in \sylcipher/ in Figure~\ref{fig:ablation}.

\noindent\textbf{\sylcipher/ is robust to syllabifiers.} We first test alternative syllabifiers beyond Pyphen, including the phoneme-based Syllabify\footnote{https://github.com/kylebgorman/syllabify} and a character-based approach using byte-pair encoding (BPE) tokenization~\cite{sennrich-etal-2016-neural}. The latter is attractive because it is language-agnostic and integrated easily with spoken language models. For the BPE-based syllabifier, we train the first stage of SuperBPE~\cite{liu-etal-2025-superbpe} on LibriSpeech with a 17k vocabulary, roughly matching the number of syllable types. We then split BPE tokens containing multiple non-consecutive vowels and merge consecutive tokens to ensure each unit contains at least one vowel (details in Appendix~\ref{app:code_bpe+}). We refer to this method as BPE+. All syllabifier variants are trained only under the \emph{fixed-boundary stage}, since later stages show correlated trends. Instead of CER, we report syllable error rate (SER) to more directly reflects syllable-level performance. Among the methods, Pyphen+ achieves the lowest SER, outperforming even the phoneme-based Syllabify. As shown in Figure~\ref{tab:ablation}, BPE+ yields weaker but still competitive results, demonstrating that \sylcipher/ can generalize to linguistically simpler, resource-free syllabifiers.

\noindent\textbf{Clamp-based soft-pooler trains more stably.} In Figure~\ref{fig:ser_vs_pooler}, we test our modified soft-pooler in \eqref{eq:softpooler}, which uses a clamp-based $\sigma_{\epsilon}$ function (``Clamp''), against alternatives based on $\tanh$~\citep{Bhati2022-scpc} (``Tanh'') and softmax~\citep{wang2024unsupervised}) (``Softmax''). Following prior work, we set the tapering parameter $\epsilon=0.5$ for ``Clamp'' and $0.1$ for the other two approaches. On both SpokenCOCO and \aishell/, Clamp consistently outperforms the sigmoid-based poolers by 6-36\%. This suggests that our design provides more stable and efficient training dynamics.\liming{Showing some pooling masks and spectrograms}

\noindent\textbf{\sylcipher/ works with different vocabulary sizes.} We also experiment with varying the non-\texttt{<OOV>} vocabulary size in Figure~\ref{fig:ser_vs_token_vocab}. \sylcipher/ displays consistent SERs across a wide range of vocabulary sizes, reaching the lowest overall SER with 2048 tokens and the lowest non-\texttt{<OOV>} SER with 1024 tokens.

\secvsp
\section{Conclusion}\label{sec:concl}
\secvsp
In this work, we introduced \sylcipher/, a UASR system that avoids phoneme-level resources such as G2P by recognizing speech at the \emph{syllable level}. Under the G2P-free setting, \sylcipher/ outperforms the best existing systems by 17-40\% relative CER on LibriSpeech and shows generalizability across other domains on SpokenCOCO, narrowing the gap with G2P-based systems. It also demonstrates cross-lingual robustness: on Mandarin, a tonal language where phoneme-based methods fail to converge, \sylcipher/ achieves a PER of 13\%. These results suggest that syllable-level modeling is a viable alternative to phoneme-level UASR and can push the horizon of more accessible and inclusive spoken language technology. % Please see Appendix~\ref{app:limit} for limitations.

%\section{Limitations}\label{app:limit}
\noindent\textbf{Limitations.} \sylcipher/ is not yet language-universal, since different languages use different writing systems and require linguistic knowledge to properly syllabify. For example, languages such as Hebrew and Arabic omits vowels in their writing system, which could pose challenges to existing syllabifiers.
% Most of our results rely on the Pyphen syllabifier, which encodes thousands of linguistically informed rules despite operating at the character level. 
While our experiments show that the method can be adapted to more resource-efficient tokenizers such as BPE with minimal modifications, coming up with a language-universal tokenization method remains an open problem.
% developing UASR systems based on truly language-universal tokenization remains an important direction of future work.
Further, the iterative training procedure can be further simplified into an end-to-end approach. Lastly, improving the robustness of \sylcipher/ under domain mismatch between speech and text remains an open challenge. 

% \section{Ethics statement}
% \section{Reproducibility statement}



\bibliography{iclr2026_conference}

\begin{thebibliography}{49}
\providecommand{\natexlab}[1]{#1}
\providecommand{\url}[1]{\texttt{#1}}
\expandafter\ifx\csname urlstyle\endcsname\relax
  \providecommand{\doi}[1]{doi: #1}\else
  \providecommand{\doi}{doi: \begingroup \urlstyle{rm}\Url}\fi

\bibitem[Ao et~al.(2022)Ao, Wang, Zhou, Wang, Ren, Wu, Liu, Ko, Li, Zhang, Wei,
  Qian, Li, and Wei]{Ao2022-speecht5}
Junyi Ao, Rui Wang, Long Zhou, Chengyi Wang, Shuo Ren, Yu~Wu, Shujie Liu, Tom
  Ko, Qing Li, Yu~Zhang, Zhihua Wei, Yao Qian, Jinyu Li, and Furu Wei.
\newblock {SpeechT5}: Unified-modal encoder-decoder pre-training for spoken
  language processing.
\newblock In Smaranda Muresan, Preslav Nakov, and Aline Villavicencio (eds.),
  \emph{Proceedings of the 60th Annual Meeting of the Association for
  Computational Linguistics (Volume 1: Long Papers), {ACL} 2022, Dublin,
  Ireland, May 22-27, 2022}, pp.\  5723--5738. Association for Computational
  Linguistics, 2022.
\newblock \doi{10.18653/V1/2022.ACL-LONG.393}.
\newblock URL \url{https://doi.org/10.18653/v1/2022.acl-long.393}.

\bibitem[Arora et~al.(2025)Arora, Chang, Chien, Peng, Wu, Adi, Dupoux, Lee,
  Livescu, and Watanabe]{arora2025-slm}
Siddhant Arora, Kai-Wei Chang, Chung-Ming Chien, Yifan Peng, Haibin Wu, Yossi
  Adi, Emmanuel Dupoux, Hung-Yi Lee, Karen Livescu, and Shinji Watanabe.
\newblock On the landscape of spoken language models: A comprehensive survey.
\newblock \emph{TMLR}, 2025.

\bibitem[Artetxe et~al.(2018{\natexlab{a}})Artetxe, Labaka, and
  Agirre]{artetxe2018unsupervised}
Mikel Artetxe, Gorka Labaka, and Eneko Agirre.
\newblock Unsupervised statistical machine translation.
\newblock In \emph{Proceedings of the 2018 Conference on Empirical Methods in
  Natural Language Processing}, pp.\  3632--3642, 2018{\natexlab{a}}.

\bibitem[Artetxe et~al.(2018{\natexlab{b}})Artetxe, Labaka, Agirre, and
  Cho]{Artetxe18-unsupnmt}
Mikel Artetxe, Gorka Labaka, Eneko Agirre, and Kyunghyun Cho.
\newblock Unsupervised neural machine translation.
\newblock In \emph{6th International Conference on Learning Representations,
  {ICLR} 2018, Vancouver, BC, Canada, April 30 - May 3, 2018, Conference Track
  Proceedings}. OpenReview.net, 2018{\natexlab{b}}.
\newblock URL \url{https://openreview.net/forum?id=Sy2ogebAW}.

\bibitem[Artetxe et~al.(2019)Artetxe, Labaka, and
  Agirre]{artetxe-etal-2019-effective}
Mikel Artetxe, Gorka Labaka, and Eneko Agirre.
\newblock An effective approach to unsupervised machine translation.
\newblock In Anna Korhonen, David Traum, and Llu{\'i}s M{\`a}rquez (eds.),
  \emph{Proceedings of the 57th Annual Meeting of the Association for
  Computational Linguistics}, pp.\  194--203, Florence, Italy, July 2019.
  Association for Computational Linguistics.
\newblock \doi{10.18653/v1/P19-1019}.
\newblock URL \url{https://aclanthology.org/P19-1019/}.

\bibitem[Baade et~al.(2025)Baade, Peng, and Harwath]{Baade2025-syllablelm}
Alan Baade, Puyuan Peng, and David Harwath.
\newblock Syllablelm: Learning coarse semantic units for speech language
  models.
\newblock In \emph{ICLR}, 2025.

\bibitem[Baevski et~al.(2020)Baevski, Zhou, Mohamed, and
  Auli]{Baevski2020-wav2vec2}
Alexei Baevski, Yuhao Zhou, Abdelrahman Mohamed, and Michael Auli.
\newblock wav2vec 2.0: {A} framework for self-supervised learning of speech
  representations.
\newblock In Hugo Larochelle, Marc'Aurelio Ranzato, Raia Hadsell,
  Maria{-}Florina Balcan, and Hsuan{-}Tien Lin (eds.), \emph{Advances in Neural
  Information Processing Systems 33: Annual Conference on Neural Information
  Processing Systems 2020, NeurIPS 2020, December 6-12, 2020, virtual}, pp.\
  12449--12460, 2020.
\newblock URL
  \url{https://proceedings.neurips.cc/paper/2020/hash/92d1e1eb1cd6f9fba3227870bb6d7f07-Abstract.html}.

\bibitem[Baevski et~al.(2021)Baevski, Hsu, Conneau, and
  Auli]{Baevski2021-wav2vec-u}
Alexei Baevski, Wei{-}Ning Hsu, Alexis Conneau, and Michael Auli.
\newblock Unsupervised speech recognition.
\newblock In Marc'Aurelio Ranzato, Alina Beygelzimer, Yann~N. Dauphin, Percy
  Liang, and Jennifer~Wortman Vaughan (eds.), \emph{Advances in Neural
  Information Processing Systems 34: Annual Conference on Neural Information
  Processing Systems 2021, NeurIPS 2021, December 6-14, 2021, virtual}, pp.\
  27826--27839, 2021.
\newblock URL
  \url{https://papers.nips.cc/paper_files/paper/2021/hash/ea159dc9788ffac311592613b7f71fbb-Abstract.html}.

\bibitem[Bhati et~al.(2022)Bhati, Villalba, Zelasko, Moro{-}Vel{\'{a}}zquez,
  and Dehak]{Bhati2022-scpc}
Saurabhchand Bhati, Jes{\'{u}}s Villalba, Piotr Zelasko, Laureano
  Moro{-}Vel{\'{a}}zquez, and Najim Dehak.
\newblock Unsupervised speech segmentation and variable rate representation
  learning using segmental contrastive predictive coding.
\newblock \emph{{IEEE} {ACM} Trans. Audio Speech Lang. Process.}, 30:\penalty0
  2002--2014, 2022.
\newblock \doi{10.1109/TASLP.2022.3180684}.
\newblock URL \url{https://doi.org/10.1109/TASLP.2022.3180684}.

\bibitem[Chen et~al.(2019)Chen, Tsai, Liu, Lee, and Lee]{Chen2019-uasr}
Kuan{-}Yu Chen, Che{-}Ping Tsai, Da{-}Rong Liu, Hung{-}yi Lee, and Lin{-}Shan
  Lee.
\newblock Completely unsupervised phoneme recognition by a generative
  adversarial network harmonized with iteratively refined hidden markov models.
\newblock In Gernot Kubin and Zdravko Kacic (eds.), \emph{Interspeech 2019,
  20th Annual Conference of the International Speech Communication Association,
  Graz, Austria, 15-19 September 2019}, pp.\  1856--1860. {ISCA}, 2019.
\newblock \doi{10.21437/INTERSPEECH.2019-2068}.
\newblock URL \url{https://doi.org/10.21437/Interspeech.2019-2068}.

\bibitem[Chen et~al.(2022)Chen, Wang, Chen, Wu, Liu, Chen, Li, Kanda, Yoshioka,
  Xiao, Wu, Zhou, Ren, Qian, Qian, Wu, Zeng, Yu, and Wei]{chen2022-wavlm}
Sanyuan Chen, Chengyi Wang, Zhengyang Chen, Yu~Wu, Shujie Liu, Zhuo Chen, Jinyu
  Li, Naoyuki Kanda, Takuya Yoshioka, Xiong Xiao, Jian Wu, Long Zhou, Shuo Ren,
  Yanmin Qian, Yao Qian, Jian Wu, Michael Zeng, Xiangzhan Yu, and Furu Wei.
\newblock {WavLM}: Large-scale self-supervised pre-training for full stack
  speech processing.
\newblock \emph{{IEEE} J. Sel. Top. Signal Process.}, 16\penalty0 (6):\penalty0
  1505--1518, 2022.
\newblock \doi{10.1109/JSTSP.2022.3188113}.
\newblock URL \url{https://doi.org/10.1109/JSTSP.2022.3188113}.

\bibitem[Chen et~al.(2024)Chen, Zhang, Peng, Li, Tian, Shi, Chang, Maiti,
  Livescu, and Watanabe]{chen2024-xeus}
William Chen, Wangyou Zhang, Yifan Peng, Xinjian Li, Jinchuan Tian, Jiatong
  Shi, Xuankai Chang, Soumi Maiti, Karen Livescu, and Shinji Watanabe.
\newblock Towards robust speech representation learning for thousands of
  languages.
\newblock In \emph{Proceedings of the 2024 Conference on Empirical Methods in
  Natural Language Processing (EMNLP)}, pp.\  10205--10224. Association for
  Computational Linguistics, 2024.
\newblock \doi{10.48550/arXiv.2407.00837}.
\newblock Preprint available on arXiv: arXiv:2407.00837.

\bibitem[Cho et~al.(2024)Cho, Mohamed, Li, Black, and
  Anumanchipalli]{cho2023-sdhubert}
Cheol~Jun Cho, Abdelrahman Mohamed, Shang-Wen Li, Alan~W Black, and Gopala~K
  Anumanchipalli.
\newblock Sd-hubert: Sentence-level self-distillation induces syllabic
  organization in hubert.
\newblock In \emph{ICASSP}, 2024.

\bibitem[Cho et~al.(2025)Cho, Lee, Gupta, Agarwal, Chen, Black, and
  Anumanchipalli]{Cho2025-sylber}
Cheol~Jun Cho, Nicholas Lee, Akshat Gupta, Dhruv Agarwal, Ethan Chen, Alan~W.
  Black, and Gopala~K. Anumanchipalli.
\newblock Sylber: Syllabic embedding representation of speech from raw audio.
\newblock In \emph{ICLR}, Singapore, 2025.

\bibitem[Chu et~al.(2024)Chu, Xu, Yang, Wei, Wei, Guo, Leng, Lv, He, Lin,
  et~al.]{chu2024qwen2}
Yunfei Chu, Jin Xu, Qian Yang, Haojie Wei, Xipin Wei, Zhifang Guo, Yichong
  Leng, Yuanjun Lv, Jinzheng He, Junyang Lin, et~al.
\newblock Qwen2-audio technical report.
\newblock \emph{arXiv preprint arXiv:2407.10759}, 2024.

\bibitem[Chung et~al.(2021)Chung, Zhang, Han, Chiu, Qin, Pang, and
  Wu]{chung2021w2vbert}
Yu-An Chung, Yu~Zhang, Wei Han, Chung-Cheng Chiu, James Qin, Ruoming Pang, and
  Yonghui Wu.
\newblock {W2v-BERT}: Combining contrastive learning and masked language
  modeling for self-supervised speech pre-training.
\newblock In \emph{2021 IEEE Automatic Speech Recognition and Understanding
  Workshop (ASRU)}, pp.\  244--250. IEEE, 2021.

\bibitem[Conneau et~al.(2021)Conneau, Baevski, Collobert, Mohamed, and
  Auli]{conneau21_interspeech}
Alexis Conneau, Alexei Baevski, Ronan Collobert, Abdelrahman Mohamed, and
  Michael Auli.
\newblock Unsupervised cross-lingual representation learning for speech
  recognition.
\newblock In \emph{Interspeech 2021}, pp.\  2426--2430, 2021.
\newblock \doi{10.21437/Interspeech.2021-329}.

\bibitem[D{\'e}fossez et~al.(2024)D{\'e}fossez, Mazar{\'e}, Orsini, Royer,
  P{\'e}rez, J{\'e}gou, Grave, and Zeghidour]{defossez2024moshi}
Alexandre D{\'e}fossez, Laurent Mazar{\'e}, Manu Orsini, Am{\'e}lie Royer,
  Patrick P{\'e}rez, Herv{\'e} J{\'e}gou, Edouard Grave, and Neil Zeghidour.
\newblock Moshi: a speech-text foundation model for real-time dialogue.
\newblock \emph{arXiv preprint arXiv:2410.00037}, 2024.

\bibitem[Fuchs \& Hoshen(2023)Fuchs and Hoshen]{fuchs2023-gradseg}
Tzeviya~Sylvia Fuchs and Yedid Hoshen.
\newblock Unsupervised word segmentation using temporal gradient pseudo-labels.
\newblock In \emph{{IEEE} International Conference on Acoustics, Speech and
  Signal Processing {ICASSP} 2023, Rhodes Island, Greece, June 4-10, 2023},
  pp.\  1--5. {IEEE}, 2023.
\newblock \doi{10.1109/ICASSP49357.2023.10095363}.
\newblock URL \url{https://doi.org/10.1109/ICASSP49357.2023.10095363}.

\bibitem[Glass(2012)]{Glass2012-unsup-speech}
James Glass.
\newblock Towards unsupervised speech processing.
\newblock In \emph{International Conference on Information Sciences, Signal
  Processing and their Applications}, 2012.
\newblock URL
  \url{https://groups.csail.mit.edu/sls/publications/2012/Glass-ISSPA12.pdf}.

\bibitem[Hoshen \& Wolf(2018)Hoshen and Wolf]{hoshen2018non}
Yedid Hoshen and Lior Wolf.
\newblock Non-adversarial unsupervised word translation.
\newblock In \emph{Proceedings of the 2018 Conference on Empirical Methods in
  Natural Language Processing}, pp.\  469--478, 2018.

\bibitem[Hsu et~al.(2021{\natexlab{a}})Hsu, Bolte, Tsai, Lakhotia,
  Salakhutdinov, and Mohamed]{Hsu2022-hubert}
Wei{-}Ning Hsu, Benjamin Bolte, Yao{-}Hung~Hubert Tsai, Kushal Lakhotia, Ruslan
  Salakhutdinov, and Abdelrahman Mohamed.
\newblock {HuBERT}: Self-supervised speech representation learning by masked
  prediction of hidden units.
\newblock \emph{{IEEE} {ACM} Trans. Audio Speech Lang. Process.}, 29:\penalty0
  3451--3460, 2021{\natexlab{a}}.
\newblock \doi{10.1109/TASLP.2021.3122291}.
\newblock URL \url{https://doi.org/10.1109/TASLP.2021.3122291}.

\bibitem[Hsu et~al.(2021{\natexlab{b}})Hsu, Harwath, Song, and
  Glass]{hsu2021textfree}
Wei-Ning Hsu, David Harwath, Chunxi Song, and James Glass.
\newblock Text-free image-to-speech synthesis using learned segmental units.
\newblock In \emph{Proceedings of the 59th Annual Meeting of the Association
  for Computational Linguistics (ACL)}, pp.\  5820--5831, 2021{\natexlab{b}}.
\newblock \doi{10.18653/v1/2021.acl-long.454}.

\bibitem[{Kahn} et~al.(2020){Kahn}, {Rivière}, {Zheng}, {Kharitonov}, {Xu},
  {Mazaré}, {Karadayi}, {Liptchinsky}, {Collobert}, {Fuegen}, {Likhomanenko},
  {Synnaeve}, {Joulin}, {Mohamed}, and {Dupoux}]{librilight}
J.~{Kahn}, M.~{Rivière}, W.~{Zheng}, E.~{Kharitonov}, Q.~{Xu}, P.~E.
  {Mazaré}, J.~{Karadayi}, V.~{Liptchinsky}, R.~{Collobert}, C.~{Fuegen},
  T.~{Likhomanenko}, G.~{Synnaeve}, A.~{Joulin}, A.~{Mohamed}, and E.~{Dupoux}.
\newblock Libri-light: A benchmark for asr with limited or no supervision.
\newblock In \emph{ICASSP}, pp.\  7669--7673, 2020.

\bibitem[Lample et~al.(2018{\natexlab{a}})Lample, Conneau, Denoyer, and
  Ranzato]{Lample2018-unsupmtmono}
Guillaume Lample, Alexis Conneau, Ludovic Denoyer, and Marc'Aurelio Ranzato.
\newblock Unsupervised machine translation using monolingual corpora only.
\newblock In \emph{6th International Conference on Learning Representations,
  {ICLR} 2018, Vancouver, BC, Canada, April 30 - May 3, 2018, Conference Track
  Proceedings}. OpenReview.net, 2018{\natexlab{a}}.
\newblock URL \url{https://openreview.net/forum?id=rkYTTf-AZ}.

\bibitem[Lample et~al.(2018{\natexlab{b}})Lample, Ott, Conneau, Denoyer, and
  Ranzato]{lample2018phrase}
Guillaume Lample, Myle Ott, Alexis Conneau, Ludovic Denoyer, and Marc’Aurelio
  Ranzato.
\newblock Phrase-based \& neural unsupervised machine translation.
\newblock In \emph{Proceedings of the 2018 Conference on Empirical Methods in
  Natural Language Processing}, pp.\  5039--5049, 2018{\natexlab{b}}.

\bibitem[Liu et~al.(2022)Liu, Lai, Hsu, Auli, Baevski, and Glass]{Liu2022-utts}
Alexander~H. Liu, Cheng-I~Jeff Lai, Wei-Ning Hsu, Michael Auli, Alexei Baevski,
  and James Glass.
\newblock Simple and effective unsupervised speech synthesis.
\newblock In \emph{Interspeech 2022}, pp.\  843--847. ISCA, 2022.
\newblock \doi{10.21437/Interspeech.2022-11071}.

\bibitem[Liu et~al.(2023)Liu, Hsu, Auli, and Baevski]{Liu2023-wav2vecu2}
Alexander~H. Liu, Wei-Ning Hsu, Michael Auli, and Alexei Baevski.
\newblock Towards end-to-end unsupervised speech recognition.
\newblock In \emph{2022 IEEE Spoken Language Technology Workshop (SLT)}, pp.\
  221--228, 2023.
\newblock \doi{10.1109/SLT54892.2023.10023187}.

\bibitem[Liu et~al.(2025)Liu, Hayase, Hofmann, Oh, Smith, and
  Choi]{liu-etal-2025-superbpe}
Alisa Liu, Jonathan Hayase, Valentin Hofmann, Sewoong Oh, Noah~A Smith, and
  Yejin Choi.
\newblock {SuperBPE}: Space travel for language models.
\newblock In \emph{Second Conference on Language Modeling}, 2025.
\newblock URL \url{https://arxiv.org/abs/2503.13423}.

\bibitem[Liu et~al.(2018)Liu, Chen, Lee, and shan Lee]{Liu2018-asru}
Da-Rong Liu, Kuan-Yu Chen, Hung-Yi Lee, and Lin shan Lee.
\newblock Completely unsupervised phoneme recognition by adversarially learning
  mapping relationships from audio embeddings.
\newblock In \emph{Interspeech}, 2018.
\newblock URL
  \url{https://www.isca-speech.org/archive_v0/Interspeech_2018/pdfs/1800.pdf}.

\bibitem[Ma et~al.(2019)Ma, McDuff, and Song]{ma2019-unpaired}
Shuang Ma, Daniel McDuff, and Yale Song.
\newblock Unpaired image-to-speech synthesis with multimodal information
  bottleneck.
\newblock In \emph{Proceedings of the IEEE/CVF International Conference on
  Computer Vision (ICCV)}, pp.\  7515--7524, 2019.
\newblock \doi{10.1109/ICCV.2019.00765}.
\newblock URL
  \url{https://openaccess.thecvf.com/content_ICCV_2019/html/Ma_Unpaired_Image-to-Speech_Synthesis_With_Multimodal_Information_Bottleneck_ICCV_2019_paper.html}.

\bibitem[Mohamed et~al.(2022)Mohamed, yi~Lee, Borgholt, Havtorn, Edin, Igel,
  Kirchhoff, Li, Livescu, Maal{\o}e, et~al.]{mohamed2022self}
Abdelrahman Mohamed, Hung yi~Lee, Lasse Borgholt, Jakob~D. Havtorn, Joakim
  Edin, Christian Igel, Katrin Kirchhoff, Shang-Wen Li, Karen Livescu, Lars
  Maal{\o}e, et~al.
\newblock Self-supervised speech representation learning: A review.
\newblock \emph{IEEE Journal of Selected Topics in Signal Processing},
  16\penalty0 (6):\penalty0 1179--1210, 2022.
\newblock \doi{10.1109/JSTSP.2022.3203489}.

\bibitem[Ni et~al.(2022)Ni, Wang, Gao, Qian, Zhang, Chang, and
  Hasegawa{-}Johnson]{Ni-unsuptts-interspeech2022}
Junrui Ni, Liming Wang, Heting Gao, Kaizhi Qian, Yang Zhang, Shiyu Chang, and
  Mark Hasegawa{-}Johnson.
\newblock Unsupervised text-to-speech synthesis by unsupervised automatic
  speech recognition.
\newblock In Hanseok Ko and John H.~L. Hansen (eds.), \emph{Interspeech 2022,
  23rd Annual Conference of the International Speech Communication Association,
  Incheon, Korea, 18-22 September 2022}, pp.\  461--465. {ISCA}, 2022.
\newblock \doi{10.21437/INTERSPEECH.2022-816}.
\newblock URL \url{https://doi.org/10.21437/Interspeech.2022-816}.

\bibitem[Ni et~al.(2025)Ni, Wang, Zhang, Qian, Gao, Hasegawa-Johnson, and
  Yoo]{Ni2025-jstti}
Junrui Ni, Liming Wang, Yang Zhang, Kaizhi Qian, Heting Gao, Mark
  Hasegawa-Johnson, and Chang~D. Yoo.
\newblock Towards unsupervised speech recognition without pronunciation models.
\newblock \emph{arXiv}, 2025.
\newblock URL \url{https://arxiv.org/pdf/2406.08380}.

\bibitem[Panayotov et~al.(2015{\natexlab{a}})Panayotov, Chen, Povey, and
  Khudanpur]{Panayotov15-LibriSpeech}
Vassil Panayotov, Guoguo Chen, Daniel Povey, and Sanjeev Khudanpur.
\newblock Librispeech: An {ASR} corpus based on public domain audio books.
\newblock In \emph{ICASSP}, pp.\  5206--5210, 2015{\natexlab{a}}.
\newblock \doi{10.1109/ICASSP.2015.7178964}.
\newblock URL \url{https://doi.org/10.1109/ICASSP.2015.7178964}.

\bibitem[Panayotov et~al.(2015{\natexlab{b}})Panayotov, Chen, Povey, and
  Khudanpur]{panayotov2015librispeech}
Vassil Panayotov, Guoguo Chen, Daniel Povey, and Sanjeev Khudanpur.
\newblock Librispeech: An asr corpus based on public domain audio books.
\newblock In \emph{2015 IEEE International Conference on Acoustics, Speech and
  Signal Processing (ICASSP)}, pp.\  5206--5210. IEEE, 2015{\natexlab{b}}.
\newblock \doi{10.1109/ICASSP.2015.7178964}.

\bibitem[Peng \& Harwath(2022)Peng and Harwath]{peng2022-vghubert}
Puyuan Peng and David Harwath.
\newblock Word discovery in visually grounded, self-supervised speech models.
\newblock In Hanseok Ko and John H.~L. Hansen (eds.), \emph{Interspeech 2022,
  23rd Annual Conference of the International Speech Communication Association,
  Incheon, Korea, 18-22 September 2022}, pp.\  2823--2827. {ISCA}, 2022.
\newblock \doi{10.21437/INTERSPEECH.2022-10652}.
\newblock URL \url{https://doi.org/10.21437/Interspeech.2022-10652}.

\bibitem[Peng et~al.(2023)Peng, Li, Räsänen, Mohamed, and
  Harwath]{Peng2023-syllable}
Puyuan Peng, Shang-Wen Li, Okko Räsänen, Abdelrahman Mohamed, and David
  Harwath.
\newblock Syllable segmentation and cross-lingual generalization in a visually
  grounded, self-supervised speech model.
\newblock In \emph{Interspeech}, 2023.

\bibitem[Sennrich et~al.(2016)Sennrich, Haddow, and
  Birch]{sennrich-etal-2016-neural}
Rico Sennrich, Barry Haddow, and Alexandra Birch.
\newblock Neural machine translation of rare words with subword units.
\newblock In Katrin Erk and Noah~A. Smith (eds.), \emph{Proceedings of the 54th
  Annual Meeting of the Association for Computational Linguistics (Volume 1:
  Long Papers)}, pp.\  1715--1725, Berlin, Germany, August 2016. Association
  for Computational Linguistics.
\newblock \doi{10.18653/v1/P16-1162}.
\newblock URL \url{https://aclanthology.org/P16-1162/}.

\bibitem[Shi \& Malik(1997)Shi and Malik]{shi1997normalized}
Jianbo Shi and Jitendra Malik.
\newblock Normalized cuts and image segmentation.
\newblock In \emph{Proceedings of the IEEE Conference on Computer Vision and
  Pattern Recognition (CVPR)}, pp.\  731--737, 1997.
\newblock \doi{10.1109/CVPR.1997.609407}.

\bibitem[Shi et~al.(2023)Shi, Hsu, Chung, Gao, Garcia, Watanabe, Lee, and
  yi~Lee]{shi2023-unsupslu}
Jiatong Shi, Chan-Jan Hsu, Holam Chung, Dongji Gao, Paola Garcia, Shinji
  Watanabe, Ann Lee, and Hung yi~Lee.
\newblock Bridging speech and text pre-trained models with unsupervised asr.
\newblock In \emph{Proceedings of the 2023 IEEE International Conference on
  Acoustics, Speech and Signal Processing (ICASSP)}, pp.\  1--5. IEEE, 2023.
\newblock \doi{10.1109/ICASSP49357.2023.10096827}.

\bibitem[Shi et~al.(2021)Shi, Bu, Xu, Zhang, and Li]{shi21c_interspeech}
Yao Shi, Hui Bu, Xin Xu, Shaoji Zhang, and Ming Li.
\newblock Aishell-3: A multi-speaker mandarin tts corpus.
\newblock In \emph{Interspeech 2021}, pp.\  2756--2760, 2021.
\newblock \doi{10.21437/Interspeech.2021-755}.

\bibitem[Tseng et~al.(2024)Tseng, Hu, Chiang, Tseng, Lee, Lee, and
  Sun]{Tseng2024-reborn}
Liang{-}Hsuan Tseng, En{-}Pei Hu, David~Cheng{-}Han Chiang, Yuan Tseng,
  Hung{-}yi Lee, Lin{-}Shan Lee, and Shao{-}Hua Sun.
\newblock {REBORN:} reinforcement-learned boundary segmentation with iterative
  training for unsupervised {ASR}.
\newblock \emph{CoRR}, abs/2402.03988, 2024.
\newblock \doi{10.48550/ARXIV.2402.03988}.
\newblock URL \url{https://doi.org/10.48550/arXiv.2402.03988}.

\bibitem[Vaswani et~al.(2017)]{Vaswani2017}
Vaswani et~al.
\newblock Attention is all you need.
\newblock In \emph{Neural Information Processing Systems}, pp.\  6000–6010,
  2017.

\bibitem[Wang et~al.(2023{\natexlab{a}})Wang, Inaguma, Chen, Kulikov, Tang,
  Hsu, Auli, and Pino]{wang-etal-2023-simple}
Changhan Wang, Hirofumi Inaguma, Peng-Jen Chen, Ilia Kulikov, Yun Tang,
  Wei-Ning Hsu, Michael Auli, and Juan Pino.
\newblock Simple and effective unsupervised speech translation.
\newblock In \emph{Proceedings of the 61st Annual Meeting of the Association
  for Computational Linguistics (Volume 1: Long Papers)}, pp.\  10771--10784,
  Toronto, Canada, July 2023{\natexlab{a}}. Association for Computational
  Linguistics.
\newblock \doi{10.18653/v1/2023.acl-long.602}.
\newblock URL \url{https://aclanthology.org/2023.acl-long.602/}.

\bibitem[Wang et~al.(2023{\natexlab{b}})Wang, Hasegawa-Johnson, and
  Yoo]{wang-etal-2023-unsupasr-theory}
Liming Wang, Mark Hasegawa-Johnson, and Chang Yoo.
\newblock A theory of unsupervised speech recognition.
\newblock In \emph{Proceedings of the 61st Annual Meeting of the Association
  for Computational Linguistics (Volume 1: Long Papers)}, pp.\  1192--1215,
  Toronto, Canada, July 2023{\natexlab{b}}. Association for Computational
  Linguistics.

\bibitem[Wang et~al.(2023{\natexlab{c}})Wang, Ni, Gao, Li, Chang, Fan, Wu,
  Hasegawa{-}Johnson, and Yoo]{wang-etal-2023-unsup-speech2sign}
Liming Wang, Junrui Ni, Heting Gao, Jialu Li, Kai~Chieh Chang, Xulin Fan,
  Junkai Wu, Mark Hasegawa{-}Johnson, and Chang~Dong Yoo.
\newblock Listen, decipher and sign: Toward unsupervised speech-to-sign
  language recognition.
\newblock In Anna Rogers, Jordan~L. Boyd{-}Graber, and Naoaki Okazaki (eds.),
  \emph{Findings of the Association for Computational Linguistics: {ACL} 2023,
  Toronto, Canada, July 9-14, 2023}, pp.\  6785--6800. Association for
  Computational Linguistics, 2023{\natexlab{c}}.
\newblock \doi{10.18653/V1/2023.FINDINGS-ACL.424}.
\newblock URL \url{https://doi.org/10.18653/v1/2023.findings-acl.424}.

\bibitem[Wang et~al.(2024)Wang, Hasegawa{-}Johnson, and
  Yoo]{wang2024unsupervised}
Liming Wang, Mark Hasegawa{-}Johnson, and Chang~D. Yoo.
\newblock Unsupervised speech recognition with n-skipgram and positional
  unigram matching.
\newblock In \emph{{IEEE} International Conference on Acoustics, Speech and
  Signal Processing, {ICASSP} 2024, Seoul, Republic of Korea, April 14-19,
  2024}, pp.\  10936--10940. {IEEE}, 2024.
\newblock \doi{10.1109/ICASSP48485.2024.10446327}.
\newblock URL \url{https://doi.org/10.1109/ICASSP48485.2024.10446327}.

\bibitem[Zhang \& Glass(2009)Zhang and Glass]{Zhang2009-syllable}
Yaodong Zhang and James~R. Glass.
\newblock Speech rhythm guided syllable nuclei detection.
\newblock In \emph{ICASSP}, pp.\  3797--3800, 2009.
\newblock \doi{10.1109/ICASSP.2009.4960454}.

\end{thebibliography}
\bibliographystyle{iclr2026_conference}
\newpage

\appendix
\section{Limitations}\label{app:limit}
\sylcipher/ is not yet language-universal, since different languages use different writing systems and require linguistic knowledge to properly syllabify. For example, languages such as Hebrew and Arabic omits vowels in their writing system, which could pose challenges to existing syllabifiers.
% Most of our results rely on the Pyphen syllabifier, which encodes thousands of linguistically informed rules despite operating at the character level. 
While our experiments show that the method can be adapted to more resource-efficient tokenizers such as BPE with minimal modifications, coming up with a language-universal tokenization method remains an open problem.
% developing UASR systems based on truly language-universal tokenization remains an important direction of future work.
Further, the iterative training procedure can be further simplified into an end-to-end approach. Lastly, improving the robustness of \sylcipher/ under domain mismatch between speech and text remains an open challenge. 
% Our method is also not yet fully language-universal and requires more research to extend to low-resource languages. % 

\section{Proof of Theorem~\ref{thm:main}}\label{app:proof_of_main}
We state the full version of  Theorem~\ref{thm:main} here.
\begin{theorem}\label{thm:main_re}
    Let $(f_{\tilde{X}}^*, g_{\tilde{X}}^*, f_Y^*, g_Y^*)$ be a minimizer of \eqref{eq:regularized_distribution_matching}, and suppose the following assumptions hold:
    \begin{enumerate}
        \item The speech feature sequence $X$ is a sequence of one-hot vectors with $|\gX|=|\tilde{\gX}|=|\gY|$; 
        \item The true syllable boundaries are used, i.e., $b_t(X)=\mathbbm{1}[y^*(X_t)\neq y^*(X_{t+1})]$;
        \item The tokenizer $c$ is an optimal \km/ quantizer with Euclidean distance metric and cluster size $|\tilde{\gX}|$;
        \item The true ASR $y^*$ is decomposable, i.e., $y^*(\tX)=[y^*(\tX_1),\cdots,y^*(\tX_L)]$;
        \item $f_{\tX}$ and $f_Y$ are decomposable;
        \item Assumption 1-2 in \citep{wang-etal-2023-unsupasr-theory} hold for $(\tilde{X}, Y)$.
    \end{enumerate}
    Then $f_{\tX}^*$ and $f_Y^*$ are invertible and  $q_{Y|X}^*(y|x):=\mathbbm{1}[f_Y^{*-1}\circ f_{\tilde{X}}^*\circ c \circ m(x)=y]$ satisfies $$\KL(p_Y||\E_X[q_{Y|X}^*])=0,\quad y^*(x)=\argmax_{y\in \gY^L} q^*_{Y|X}(y|x),\quad\forall x\in\gX^T.$$
\end{theorem}

The proof relies on the following lemma proven in Appendix~\ref{app:proof_of_lemma_invertible}.
\begin{lemma}\label{lemma:invertible}
Suppose Assumption 5 of Theorem~\ref{thm:main} and  Assumption 1 and 2 of \citep{wang-etal-2023-unsupasr-theory} hold for $(\tX,Y)$, and functions $g_X:\tilde{\gX}^L\mapsto [0,1],g_Y:\tilde{\gY}^L\mapsto[0,1]$ of the form in \eqref{eq:postnet} be such that $p_X=g_X\circ f_X$ and $p_Y=g_Y\circ f_Y$. Then $f_X$ and $f_Y$ are both invertible.
\end{lemma}

Now we are ready to prove the main theorem. 
\begin{proof}
The proof consists of two main parts: (i) First, we prove that at least one  minimizer $(f_{\tilde{X},1}, g_{\tilde{X},1}, f_{Y,1}, g_{Y,1})$ can achieve a minimum of 0 for \eqref{eq:regularized_distribution_matching}; (ii) then we establish that for any minimizers $(f_{\tilde{X}}^*, g_{\tilde{X}}^*, f_Y^*, g_Y^*)$, the marginal distribution of the text posterior $q_{Y|X}^*$ matches the true text distribution $p_Y$. As a result, by Assumption 4, Theorem 1 of \citep{wang-etal-2023-unsupasr-theory} guarantees that $\argmax_{y\in\gY^L}q_{Y|X}^*(y|X)$ achieves zero-error UASR. 

% Achievability
To prove (i), by the definition of $y^*$ and Assumption 1, $y^*$ is a invertible mapping between $\gX$ and $\gY$. Then let $t_i=\min\{\tau:\sum_{t=1}^{\tau}b_t(X)\geq i-1\}$ be the \emph{starting time} of each syllable and apply Assumption 2, we have
\begin{align*}
    X_s = X_t,\,\forall t_i\leq s<t< t_{i+1},\,1\leq i\leq L.
\end{align*}
As a result, the output of the soft-pooler is simply $m_i(X)=X_{t_i}.$ Further, by Assumption 3, $c$ is optimal, and we claim that this is achievable if and only if for any $(x,x')\in\gX^2,$ 
\begin{align*}
    c(x)=c(x')\Longleftrightarrow y^*(x)=y^*(x').
\end{align*}
Otherwise, suppose for some $(x,x')\in\gX^2$ such that $c(x)=c(x')=:c_0$ but $y^*(x)\neq y^*(x'),$ then for any centroid $\mu_{c_0}$ of cluster $c_0$ and triangle inequality,  
\begin{align*}
    \|x-\mu_{c(x)}\|_2+\|x'-\mu_{c(x')}\|_2=\|x-\mu_{c_0}\|_2+\|x'-\mu_{c_0}\|_2\geq \|x-x'\|_2 >0,
\end{align*}
and therefore the \km/ objective:
\begin{align*}
    \Ls_{\mathrm{km}}(\mu_1,\cdots,\mu_{|\gX|}):=\E_{X\sim p_X}\|X-\mu_{c(X)}\|^2_2>0.
\end{align*}

However, by simply setting $c(x)=y^*(x)$ for all $x\in\gX$, we have $\mu_{c(x)}=x$ for any $x\in \gX$ and $\Ls_{\mathrm{km}}=0$ due to the invertibility of $y^*.$ This contradicts with the optimality of $c$ and thus proves the ``$\Longrightarrow$'' direction. Further, this implies that the cluster size is at least $|\gY|$ and since we set the cluster size to $|\gY|$, $c\circ m$ is equivalent to $y^*$ up to a permutation. This then proves the other direction. Let $c\circ m=\pi\circ y^*$ where $\pi$ is a permutation, then the syllable-level features $\tilde{X}=\pi\circ y^*(X)$ and thus by the permutation-invariance of discrete entropy, 
\begin{align*}
    H(\tilde{X})=H(y^*(X))=H(Y).
\end{align*}
Therefore, set $f_{\tilde{X},1}$ to be a permutation on $\tilde{\gX}$, $f_{Y,1}=f_{\tilde{X},1}\circ\pi$, $g_{\tilde{X},1}=p_{\tilde{X}}\circ f_{X,1}^{-1}$ and $g_{Y,1}=p_Y\circ f_{Y,1}^{-1},$ we have
\begin{align*}
    \KL(p_{\tilde{X}}||g_{\tilde{X},1}\circ f_{\tilde{X},1})+\KL(p_Y||g_{Y,1}\circ f_{Y,1})&=\KL(p_{\tilde{X}}||p_{\tilde{X}})+\KL(p_Y||p_Y)=0,\\
\end{align*}
and 
\begin{align*}
    f_{Y,1}(y(X))=f_{\tilde{X},1}\circ \pi\circ y(X)=f_{\tilde{X},1}\circ c\circ m(X)=f_{\tilde{X},1}(\tilde{X}).
\end{align*}
The latter implies that the \pdf/ of $f_{Z,1}(Z)$ satisfies
\begin{align}
p_{f_{Z,1}(Z)}=\frac{1}{2}p_{f_{\tilde{X},1}(\tilde{X})}+\frac{1}{2}p_{f_{Y,1}(Y)}=\frac{1}{2}p_{f_{\tilde{X},1}(\tilde{X})}+\frac{1}{2}p_{f_Y(y(X)),1}=p_{f_{\tilde{X},1}(\tilde{X})},\tag{*}    
\end{align}
and thus 
$$H(f_{Z,1}(Z))=H(f_{Y,1}(Y))=H(f_{X,1}(\tX))\leq H(Y).$$
Therefore, by the nonnegativity of $\KL$, $(f_{\tilde{X},1}, g_{\tilde{X},1}, f_{Y,1}, g_{Y,1})$ is an optimal solution of \eqref{eq:regularized_distribution_matching} with a minimum of $0$, which proves $(i)$. 

% Distribution matching
To prove (ii), we first use (i) to conclude that
\begin{align*}
    p_{\tX}=g_{\tX}^*\circ f_{\tX}^*,\,p_Y=g_Y^*\circ f_Y^*.
\end{align*}
Then it amounts to prove that any minimizer $(f_{\tilde{X}}^*,g_{\tilde{X}}^*,f_Y^*,g_Y^*)$ of \eqref{eq:regularized_distribution_matching} satisfies
\begin{align}
    p_{f_Z^*(Z)}=p_{f_X^*(X)}=p_{f_Y^*(Y)}.\tag{*}
\end{align}
Since if this is the case, by Lemma~\ref{lemma:invertible}, $f_{Y}^*$ is invertible and thus for any $y\in\gY^L$,
\begin{align*}
    &p_{f_Y^{*-1}\circ f_X^*(X)}(y)\overset{(a)}{=}p_{f_X^*(X)}(f_Y^*(y))\overset{(b)}{=}p_{f_Y^*(Y)}(f_Y^*(y))\overset{(c)}{=}p_Y(y)\\
    \Longrightarrow &\KL(p_Y||\E_Xq^*_{Y|X})=0,
\end{align*}
where $(b)$ uses (*) and $(a)(c)$ uses Lemma~\ref{lemma:invertible}.

To prove (*), we use the concavity of the discrete entropy $h(p):=-\sum_xp(x)\log p(x)$ and Lemma~\ref{lemma:invertible},
\begin{align*}
    H(f_Z^*(Z))&=h\brac{\frac{1}{2}p_{f_X^*(X)}+\frac{1}{2}p_{f_Y^*(Y)}}\overset{(d)}{\geq} \frac{1}{2}h(p_{f_X^*(X)})+\frac{1}{2}h(p_{f_Y^*(Y)})\\
    &\overset{(e)}{=} \frac{1}{2}(H(X)+H(Y))=H(Y),
\end{align*}
with equality if and only if $p_{f_Y^*(Y)}=p_{f_X^*(X)},$ where $(d)$ uses the concavity of the entropy function and $(e)$ uses Lemma~\ref{lemma:invertible}. This concludes the proof of $(ii).$
\end{proof}

\section{Proof of Lemma~\ref{lemma:invertible}}\label{app:proof_of_lemma_invertible}
\begin{proof}
We prove the lemma by contradiction and focus on proving the invertibility of $f_X$ since the proof is analogous for $f_Y$. Suppose otherwise $f_X$ is not invertible, then by Assumption 5, there exists $(x',x'')\in\tilde{\gX}^2$ such that $f_X(x')=f_X(x'')$ but $x'\neq x''$, then by the definition of $g_X$,
\begin{multline}
    p_{X_i}(x')=\sum_{x_{-i}\in\tilde{\gX}^{L-1}}p_X(x_1,\cdots,x_{i-1},x',x_{i+1},\cdots,x_L)\\
    =\sum_{x_{-i}\in\gX^{L-1}}\prod_{j<i}g_X(x_j|f_X(x_1),\cdots,f_X(x_{j-1}))\cdot\\
    \prod_{k\geq i}g_X(x_k|f_X(x_1),\cdots,f_X(x_{i-1}),f_X(x'),f_X(x_{i+1})\cdots,f_X(x_{k-1}))\\
    =\sum_{x_{-i}\in\gX^{L-1}}\prod_{j<i}g_X(x_j|f_X(x_1),\cdots,f_X(x_{j-1}))\cdot\\
    \prod_{k\geq i}g_X(x_k|f_X(x_1),\cdots,f_X(x_{i-1}),f_X(x''),f_X(x_{i+1})\cdots,f_X(x_{k-1}))\\
    =\sum_{x_{-i}\in\tilde{\gX}^{L-1}}p_X(x_1,\cdots,x_{i-1},x'',x_{i+1},\cdots,x_L)=p_{X_i}(x'').
\end{multline}
Therefore, the positional unigram matrix
\begin{align*}
    P^X:=\begin{bmatrix}
        p_{X_1}^\top \\
        \vdots \\
        p_{X_L}^\top
    \end{bmatrix}
\end{align*}
is column-rank-deficient. However, Theorem 1 of \citep{wang-etal-2023-unsupasr-theory} asserts that if Assumption 1 and 2 of \citep{wang-etal-2023-unsupasr-theory} holds, $P^X$ has full column-rank, which is a contradiction. Therefore, $f_X(x')\neq f_X(x'')$ if $x'\neq x''$ and $f_X$ is invertible.
\end{proof}

\section{Implementation details of \sylcipher/}\label{app:implementation}
For English experiments, \sylcipher/ uses a \hubl/-large\footnote{https://dl.fbaipublicfiles.com/hubert/hubert\_large\_ll60k.pt} model pretrained on LibriLight~\citep{librilight} as the SSL encoder in the speech syllabifier, and initialize the soft-pooler with unsupervised boundary labels from Sylber~\citep{Cho2025-sylber}. For Mandarin, we instead use XEUS\footnote{https://huggingface.co/espnet/xeus/blob/main/model/xeus\_checkpoint\_new.pth}~\citep{chen2024-xeus}, which is pretrained on Mandarin (among other languages) and significantly outperforms \hubl/. We also finetune Sylber on \aishell/ to ensure stable convergence. For the \prenet/s, shared encoder, \postnet/s, MLM parameters, and most of the optimizer hyperparameters, we follow JSTTI~\citep{Ni2025-jstti}, as modifying them did not yield consistent improvements.
\liming{Explain how the baselines are set up?}

\newpage

\section{Pseudo-code for the Pyphen+ syllabifier}\label{app:code_pyphen+}
\begin{algorithm}[H]
\caption{Merging No-Vowel Segments and Naive Syllabification}
\begin{algorithmic}[1]
\Function{merge\_novowel\_segments}{segs}
    \State new\_segs $\gets$ [ ]
    \State n\_seg $\gets$ len(segs)
    \State i $\gets 0$
    \While{$i < n\_seg$}
        \State seg $\gets$ segs[i]
        \If{seg has no vowels}
            \If{seg is the last segment}
                \State merge seg into previous segment (if any)
            \Else
                \State merge seg into the next segment
                \State $i \gets i + 1$
            \EndIf
        \Else
            \State append seg to new\_segs
        \EndIf
        \State $i \gets i + 1$
    \EndWhile
    \State \Return new\_segs
\EndFunction

\Function{naive\_syllabify}{w}
    \State w0 $\gets$ w
    \If{w ends with a silent `e' (not ``-le'')}
        \State drop the final `e' in w
    \EndIf
    \If{w ends with ``-ed'' and root does not end with `t' or `d'}
        \State drop the `e' in ``-ed'' in w
    \EndIf

    \State Apply rule to w: split \texttt{V C-CC… V} into \texttt{VC-CC…V}
    \State Apply rule to w: split \texttt{VCCV} into \texttt{VC-CV}
    \State Apply rule to w: split \texttt{VCV} into \texttt{V-CV}

    \If{modifications were made}
        \If{w0 ended with `e'}
            \State add back `e' to w
        \ElsIf{w0 ended with `ed'}
            \State restore `ed' to w
        \EndIf
    \EndIf
    \State \Return syllabified w
\EndFunction

\Function{pyphen\_plus\_syllabify}{w}
    \State syls $\gets$ []
    \State segs $\gets$ \href{https://github.com/grantjenks/python-wordsegment}{wordsegment}.segment(w)
    \For{each seg in segs}
        \State syls\_ $\gets$ pyphen.inserted(seg)
        \State syls\_ $\gets$ merge\_novowel\_segments(syls\_)
        \If{syls\_ has only one syllable \textbf{and} \href{https://github.com/mholtzscher/syllapy}{syllapy}.count(syls\_) $>$ 1} 
            \State syls\_ $\gets$ naive\_syllabify(syls\_)
        \EndIf
        \State extend syls with syls\_ 
    \EndFor
\EndFunction
\end{algorithmic}
\end{algorithm}

\section{Pseudo-code of BPE+ syllabifier}\label{app:code_bpe+}
\begin{algorithm}[H]
\caption{Splitting and Processing BPE Tokens with Vowel Constraints}
\begin{algorithmic}[1]
\Function{split\_on\_nonconsecutive\_vowels}{token}
    \State parts $\gets$ [ ]
    \State current $\gets$ [ ]
    \State vowel\_positions $\gets$ [ ]
    \For{each character $ch$ in token}
        \State append $ch$ to current
        \If{$ch \in$ VOWELS}
            \State record position of $ch$ in current
        \EndIf
        \If{two or more vowels are non-consecutive}
            \State cut before the last vowel
            \State append left substring to parts
            \State reset current and vowel\_positions accordingly
        \EndIf
    \EndFor
    \If{current is not empty}
        \State append current to parts
    \EndIf
    \State \Return parts
\EndFunction

\Function{enforce\_vowel\_constraint}{parts}
    \State remove empty parts
    \State new\_parts $\gets$ [ ]
    \State buffer $\gets$ empty string
    \For{each part $p$ in parts}
        \State buffer $\gets$ buffer + $p$
        \If{buffer contains a vowel}
            \If{next part starts with a vowel \textbf{or} ``E''}
                \State continue without breaking
            \Else
                \State append buffer to new\_parts
                \State reset buffer
            \EndIf
        \EndIf
    \EndFor
    \If{buffer not empty}
        \If{new\_parts not empty}
            \State merge buffer into the last part
        \Else
            \State append buffer as a new part
        \EndIf
    \EndIf
    \State \Return new\_parts
\EndFunction

\Function{bpe\_plus\_syllabify}{tokens}
    \State first\_split $\gets$ apply split\_on\_nonconsecutive\_vowels to each token
    \State final\_parts $\gets$ enforce\_vowel\_constraint(first\_split)
    \State \Return final\_parts
\EndFunction
\end{algorithmic}
\end{algorithm}

\section{Spectrogram examples on LibriSpeech}\label{app:spec_libri}
\begin{figure}
    \centering
    \begin{subfigure}{0.99\textwidth}
    \includegraphics[width=0.99\textwidth]{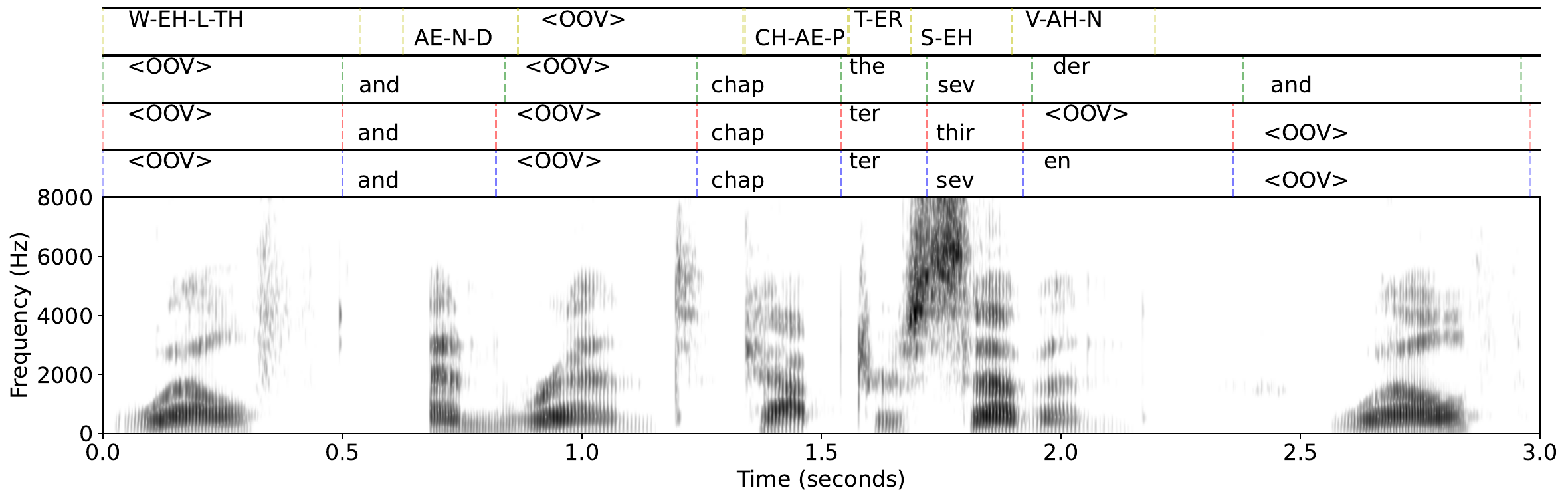}
    \caption{Reference transcript: ``Wealth and rent chap-ter sev-en wealth''.}    
    \end{subfigure}
    \begin{subfigure}{0.99\textwidth}
    \includegraphics[width=0.99\textwidth]{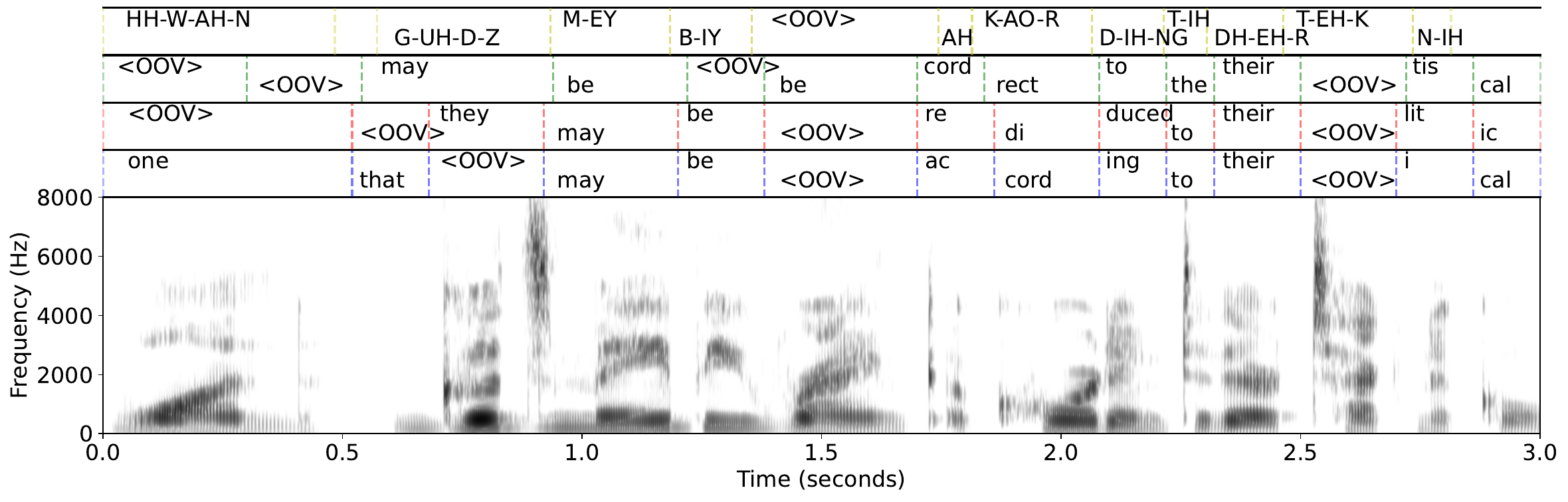}
    \caption{Reference transcript: ``One goods may be ranked ac-cord-ing to their tech-ni-cal''.}    
    \end{subfigure}
    \begin{subfigure}{0.99\textwidth}
    \includegraphics[width=0.99\textwidth]{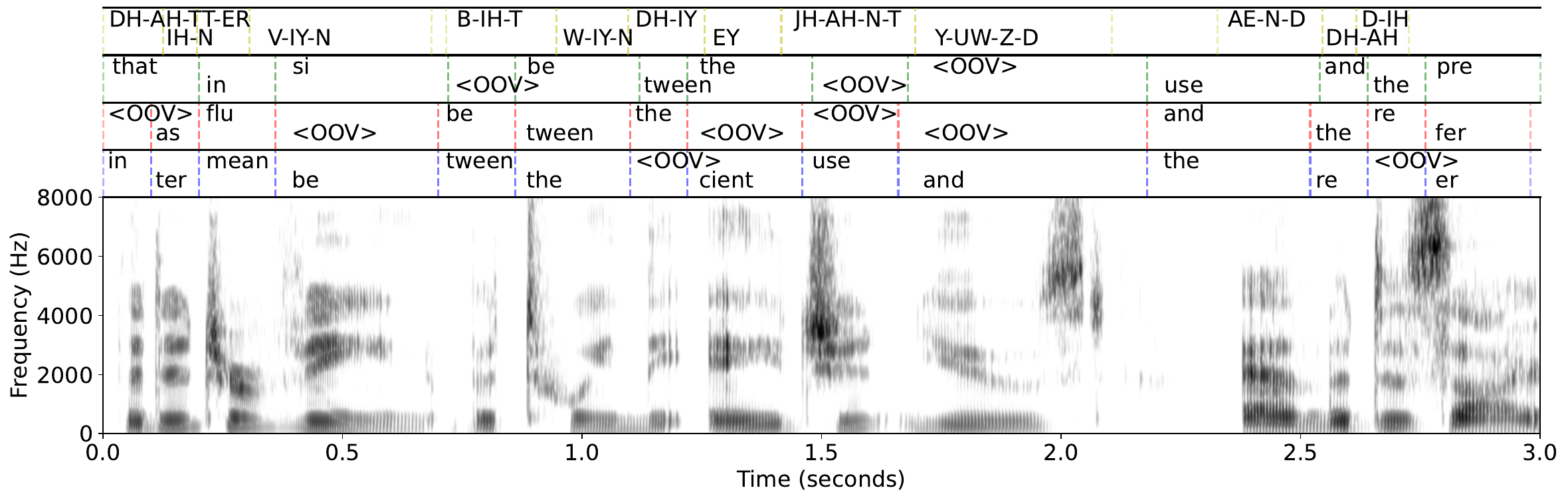}
    \caption{Reference transcript: ``That in-ter-vene be-tween the a-gent used and the de-sired''}    
    \end{subfigure}
    \begin{subfigure}{0.99\textwidth}
    \includegraphics[width=0.99\textwidth]{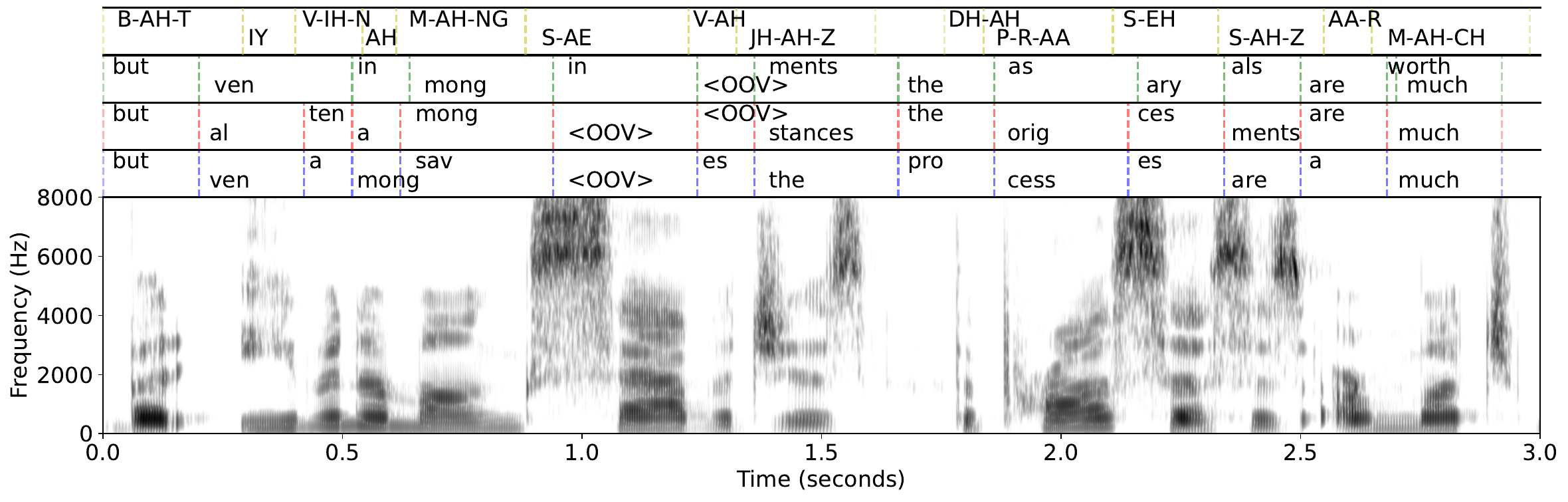}
    \caption{Reference transcript: ``But e-ven a-mong sav-ages the pro-cess-es are much''}    
    \end{subfigure}
    \caption{\textbf{Spectrograms of audio examples in our test split of LibriSpeech clean subsets (matched setting) and the predicted speech-text alignment by \sylcipher/ after different training stages.} Audios are truncated to the 3-second mark for better visualization. The alignment bars from top to bottom: Forced alignment, Sylber, Sylber+JE2E, Sylber+JE2E+PUSM.}
\end{figure}

\begin{figure}
    \centering
    \begin{subfigure}{0.99\textwidth}
    \includegraphics[width=0.99\textwidth]{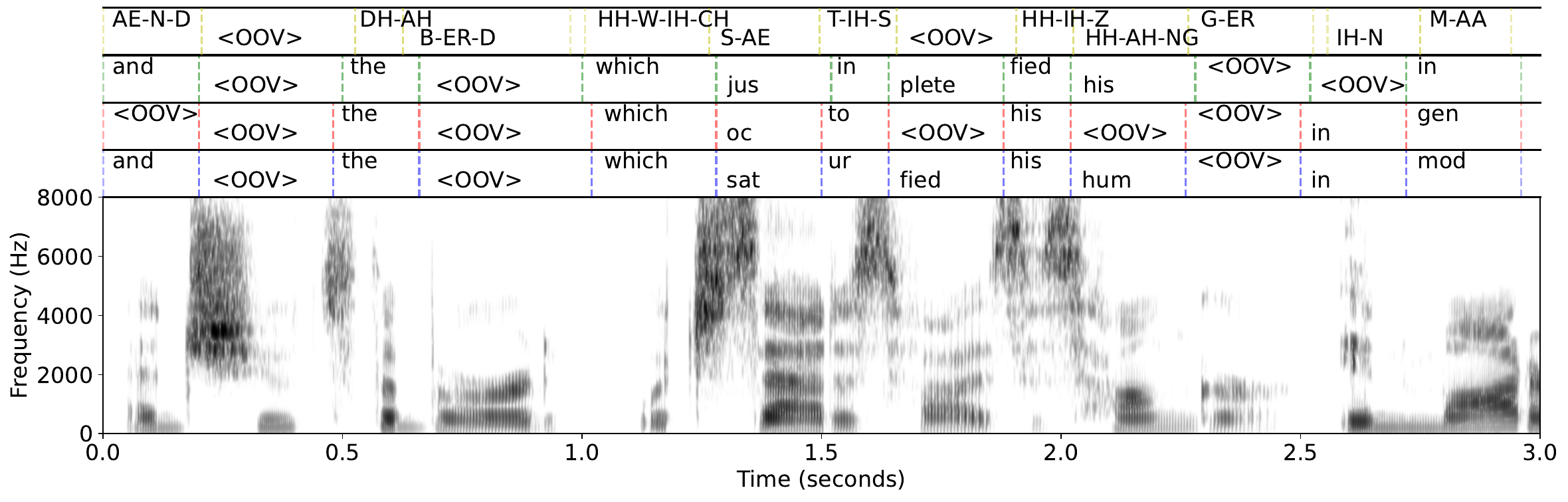}
    \caption{Reference transcript: ``And shoots the bird which sat-is-flies his hun-ger''.}    
    \end{subfigure}
    \begin{subfigure}{0.99\textwidth}
    \includegraphics[width=0.99\textwidth]{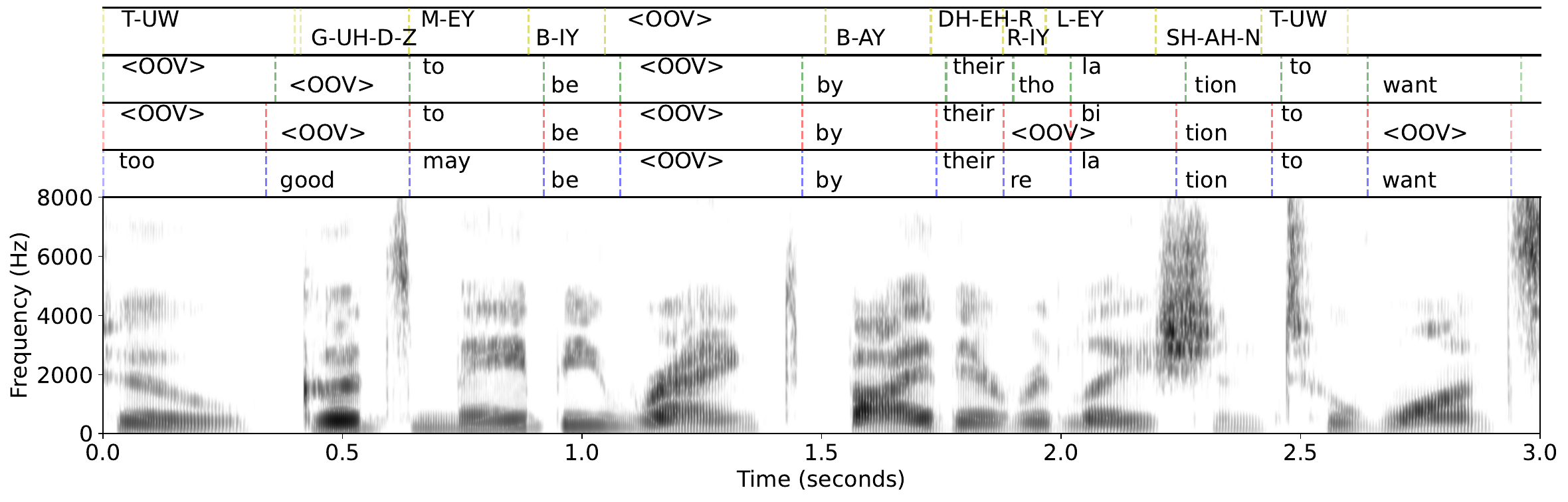}
    \caption{Reference transcript: ``two goods may be ranked by their re-la-tion to wants''.}    
    \end{subfigure}
    \begin{subfigure}{0.99\textwidth}
    \includegraphics[width=0.99\textwidth]{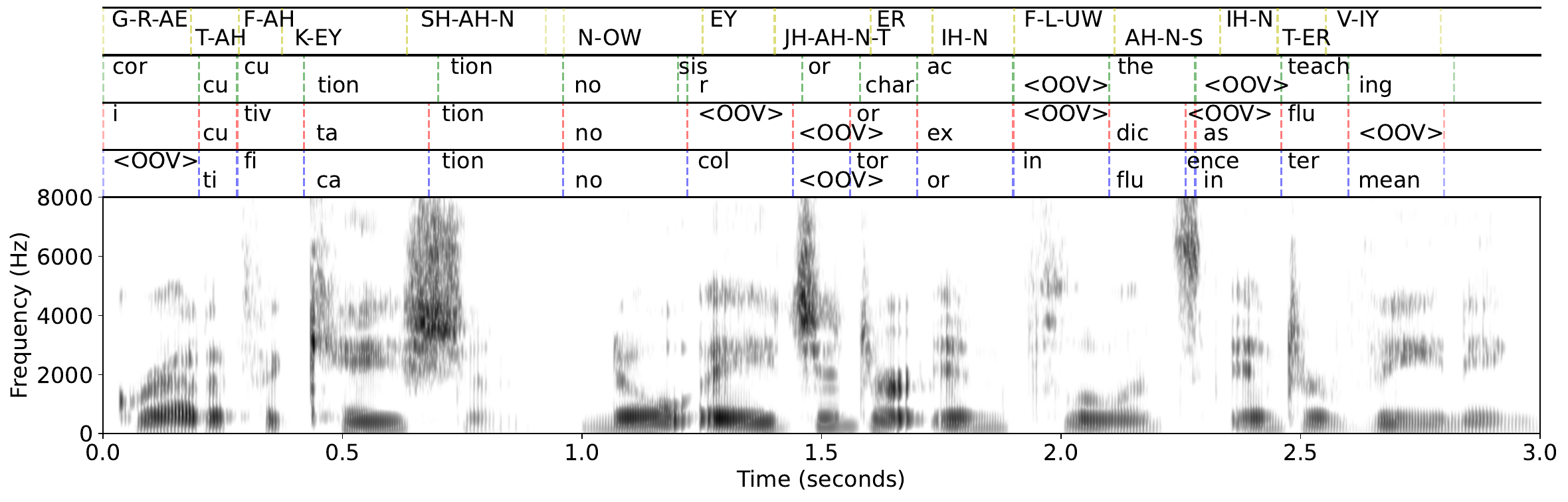}
    \caption{Reference transcript: ``Grat-i-fi-ca-tion no a-gent or in-flu-ence in-ter-ven-ing''.}    
    \end{subfigure}
    \begin{subfigure}{0.99\textwidth}
    \includegraphics[width=0.99\textwidth]{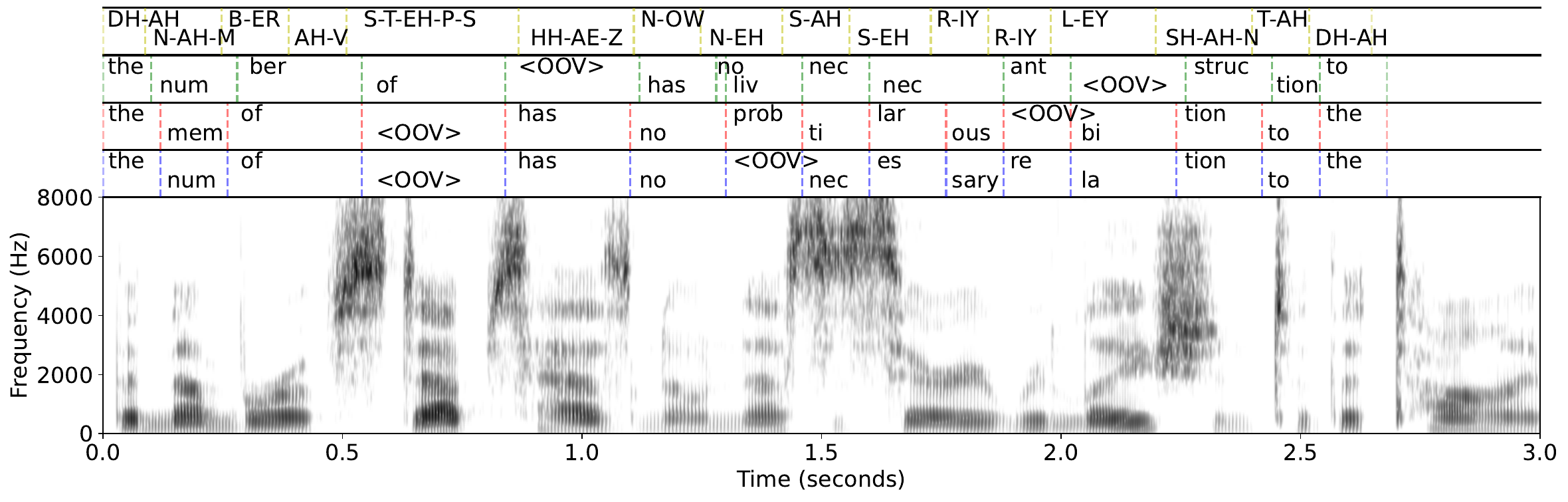}
    \caption{Reference transcript: ``The num-ber of steps has no nec-es-sary re-la-tion to the''.}    
    \end{subfigure}
    \caption{\textbf{Spectrograms of audio examples in our test split of LibriSpeech clean subsets (matched setting) and the predicted speech-text alignment by \sylcipher/ after different training stages.} Audios are truncated to the 3-second mark for better visualization. The alignment bars from top to bottom: Forced alignment, Sylber, Sylber+JE2E, Sylber+JE2E+PUSM.}
\end{figure}

\end{document}